# Tracking Point of View in Narrative

Janyce M. Wiebe*
New Mexico State University

*Third-person fictional narrative text is composed not only of passages that objectively narrate events, but also of passages that present characters' thoughts, perceptions, and inner states. Such passages take a character's* **psychological point of view**. *A language understander must determine the current psychological point of view in order to distinguish the beliefs of the characters from the facts of the story, to correctly attribute beliefs and other attitudes to their sources, and to understand the discourse relations among sentences. Tracking the psychological point of view is not a trivial problem, because many sentences are not explicitly marked for point of view, and whether the point of view of a sentence is objective or that of a character (and if the latter, which character it is) often depends on the context in which the sentence appears. Tracking the psychological point of view is the problem addressed in this work. The approach is to seek, by extensive examinations of naturally occurring narrative, regularities in the ways that authors manipulate point of view, and to develop an algorithm that tracks point of view on the basis of the regularities found. This paper presents this algorithm, gives demonstrations of an implemented system, and describes the results of some preliminary empirical studies, which lend support to the algorithm.*

## 1. Introduction

Imagine that a language understander encounters the following passage while reading a novel:

> (1)
> $^{1.1}$He [Sandy] wanted to talk to Dennys. $^{1.2}$How were they going to be able to get home from this strange desert land into which they had been cast and which was heaven knew where in all the countless solar systems in all the countless galaxies? [L'Engle, *Many Waters*, p. 91]

In this passage, the author is not objectively narrating events or describing the fictional world, but is presenting the thoughts and emotions of a character. It is to Sandy that the land is strange, and it is Sandy's uncertainty that is expressed by the question and the expression 'heaven knew where'. Unless the language understander realizes these things, it hasn't fully understood the passage.

Passages such as (1) take a character's **psychological point of view** and are composed of **subjective sentences**—sentences that present the thoughts, perceptions, and inner states of characters in the story. Notice that nothing in (1.2) explicitly specifies that the sentence is Sandy's thought. In general, only a **narrative parenthetical**, such as 'Dennys thought' in (2), serves to *explicitly* indicate both that a sentence is subjective and who its **subjective character** is.

> (2)
> Certainly, Dennys thought, anything would be better than this horrible-smelling place full of horrible little people. [L'Engle, *Many Waters*, p.

In all other cases, one must rely on less direct sources of information to determine the psychological point of view.

This paper presents an algorithm for recognizing subjective sentences and identifying their subjective characters in third-person fictional narrative text. The algorithm is based on regularities, found by extensive examination of naturally occurring text (i.e., published novels and short stories) in the ways that authors manipulate point of view. It has been implemented, and some preliminary empirical studies, which lend support to the algorithm, have also been performed.

The algorithm is described in the body of the paper, and is given in full in Appendix I. Sections 2-5 give background information and describe my approach to the problem. Sections 6 and 7 present an overview of the algorithm, specifying the input and output of the basic components, and identifying the components focused on in this work. Sections 8-10 present the bulk of the algorithm, addressing the problem of identifying subjective characters before the problem of recognizing subjective sentences. Section 11 describes the algorithm's treatment of sentences about **private-state actions**, such as sighing and looking. Sections 12-14 conclude the paper with a summary of tests of the algorithm, and discussions of the relationship between tracking point of view and anaphora resolution and of directions for future research. The algorithm is given in appendix I, demonstrations of its implementation are given in appendix II, and the results of a test of the algorithm are given in appendix III.

## 2. Point of View

### 2.1 Introduction

In face-to-face conversation, an utterance is understood with respect to the situation in which the conversation takes place (Barwise and Perry 1983). Thus, deictic expressions such as 'now', 'here', 'come', 'go', and 'just ahead' derive their meanings from the time and place of the utterance (Fillmore 1975, Lyons 1977). In fictional narrative text, however, spatial and temporal deictic terms are clearly not understood with respect to the time and place of the author's writing nor of the reader's reading. Rather, they are understood with respect to a "here" and "now" within the story (Hamburger 1973, Kuroda 1976, Banfield 1982, Bruder et al. 1986, Rapaport et al. 1989ab). Thus, the reader must track the **spatial** and **temporal** points of view with respect to which objects and events are described (Uspensky 1973).

Knowing who the speaker is is another situational component needed to understand conversation. Most obviously, the speaker is needed as the referent of first-person pronouns, but in addition she is the source of beliefs, emotions, evaluations, etc., expressed by her utterances. But in third-person narrative text, sentences can express a *character's* beliefs, emotions, etc., even when he or she is referred to in the third person. Thus, an additional deictic component, corresponding to the speaker in conversation, is needed to understand third-person narrative: the **psychological point of view** (Uspensky 1973).

Although various points of view often coincide in third-person narrative text, they need not. For example, a passage may take a character's spatial and temporal points of view without taking the psychological point of view of any character (Uspensky 1973). It is the psychological point of view with which we are concerned in this paper.

### 2.2 Subjective Sentences

Following Ann Banfield (1982), a literary theorist who analyzes point of view linguistically, we shall call sentences that take a character's psychological point of view (here-after, simply **point of view** or **POV**) **subjective**, in contrast to sentences that **objectively** narrate events or describe the fictional world. Subjective sentences present **private states** of characters—states of an experiencer holding an attitude, optionally toward an object. Varieties of private states include intellectual ones, such as someone believing, wondering, or knowing something; emotive ones, such as someone hating something or being afraid; and perceptual ones, such as someone seeing or hearing something. Thus, private states are states that are not open to objective observation or verification (Quirk et al. 1985). To refer to a private state $p$ and its components, we shall write:

$$\text{PS}\ (p, \textit{experiencer, attitude, object})$$

where *experiencer* is the person in state $p$, and *attitude* is know, believe, see, or whatever sort of private state $p$ is. Notice that "attitude" is being used as a general covering term, referring to a class of which the propositional attitudes are only a subclass.

We shall limit our scope in this paper mainly to two classes of subjective sentences, one containing those Banfield calls "represented thoughts" and "represented perceptions", and the other containing those I call "private-state reports". We will not consider **represented speech** at all (Jespersen 1924, Banfield 1982), due to its complexity, and only toward the end of the paper (section 11) will we consider a variant of private-state reports: subjective sentences about **private-state actions**.

Even restricting one's attention to the two classes mentioned above, there are many syntactic, semantic, and pragmatic features according to which one could characterize subjective sentences. Below I propose a view of subjective sentences that is useful for the specific purpose of tracking POV. For characterizations that include further and alternative distinctions, see, for example, the following works in literary theory and linguistics: Doležel 1973, Uspensky 1973, Fillmore 1974, Cohn 1978, Banfield 1982, Caenepeel 1989, Galbraith 1990, and Li 1991.

A private state is part of the meaning of any kind of subjective sentence. However, a represented thought or represented perception without a narrative parenthetical explicitly mentions only the object of the private state; the attitude and experiencer are implicit. On the other hand, a private-state report explicitly mentions the experiencer, the attitude, and, optionally, the object of a private state $p$; in addition, with the private-state report, $p$ is not the object of some other private state with an implicit experiencer and attitude. Consider the following sentences:

$^{3.1}$ Zoe was angry at Joe. $^{3.2}$ Where was he?

Sentence (3.2) is a represented thought. It mentions the object of a private state $p$ whose experiencer, Zoe, and attitude, *wonder* or a similar attitude, are implicit:

$$\begin{array}{c} \text{"Where was he?"} \\ \downarrow \\ \text{PS}(p,\ \textit{experiencer},\quad \textit{attitude},\quad \textit{object}) \end{array}$$

Sentence (3.1) mentions the experiencer, attitude, and object of a private state $p_1$:

$$\begin{array}{cccc} \text{" Zoe} & \text{was angry at} & \text{Joe."} \\ \downarrow & \downarrow & \downarrow \\ \text{PS}(p_1,\ \textit{experiencer}_1, & \textit{attitude}_1, & \textit{object}_1) \end{array}$$

Under a private-state *report* interpretation of (3.1), $p_1$ is not itself the object of some other private state.

We shall call the character whose psychological point of view is taken by a subjective sentence the **subjective character** (**SC**) of that sentence. The SC is the subject of a narrative parenthetical, when one is present; the implicit experiencer, when the sentence is a represented thought or represented perception without a narrative parenthetical; and the explicit experiencer, when the sentence is a private-state report.

In addition to private-state terms such as 'know' and 'see', subjective sentences can contain **subjective elements**, linguistic elements that express attitudes of the SC (this aspect of subjective sentences is noted in many studies of POV; the term "subjective element" is due to Banfield 1982). An example appears in the following passage (throughout this paper, sentences in cited passages are indented to reflect paragraphing in the original texts).

> (4)
> $^{4.1}$"What are you doing in here?" $^{4.2}$Suddenly she [Zoe] was furious with him [Joe].
> $^{4.3}$"Spying, of course."
> $^{4.4}$"Well of all dumb things! $^{4.5}$I thought you ran away." $^{4.6}$Joe Bunch was awful. [Oneal, *War Work*, p. 130]

The adjective "awful" in (4.6) is a subjective element, expressing Zoe's evaluation of Joe (that he is awful). Notice that (4.6) is a represented thought—*Zoe thought* is implicit—and (4.2) is a private-state report, reporting Zoe's private state of being furious with Joe.

As will be specified shortly in section 6, the output of the algorithm is the POV of a sentence, not a representation of its meaning. One interesting issue not addressed in this paper is how one might represent the difference between, for example, "John wondered where he was" and " 'Where was he?,' John wondered."

But one distinction that is crucial for tracking POV is the distinction between the interpretation of a sentence as either objective or subjective, and the kinds of states of affairs that the sentence is about. The former is a pragmatic issue, concerning the *use* of a sentence to present objective or subjective information about the fictional world (Kuroda 1976). The latter is a semantic issue, concerning the word senses of lexical items in the sentence. For instance, (5) below can be used to objectively narrate a past action or to portray a character's thought of one. In either case, the sentence is about an action.

> (5) Gus had taken them back to town.

Now consider (6):

> (6) Gus had a cornish hen.

Whether the word sense of 'had' in (6) is a state or an action (eating) is a separate question from whether the sentence is objective or subjective (similarly, Hirst (1987) suggests that lexical ambiguity is orthogonal to speech act ambiguity in conversation). As specified below in section 6.1, the type of state of affairs that each clause is about is part of the input to the algorithm.

In addition to being objective or a character's subjective sentence, a sentence may express the attitudes of an **overt narrator** (Chatman 1978). But this paper will focus only on texts without overt narrators. In addition, we will not consider passages with irony and humor that do not originate with a character, or passages that are written in the style of an epic, parable, or folktale. As well, we will not try to account for fiction that is experimental in its manipulation of POV. Finally, we will only consider narrative sentences in which the tenses are "shifted" (Jespersen 1924), that is, in which the simple past is used for the narrative present, and the past perfective is used for the narrative past. Note that it is not difficult to find published third-person texts that satisfy these criteria.

## 3. Importance of Tracking Point of View

The importance to narrative understanding of recognizing characters' intentions has long been recognized in AI (see, for example, Wilensky 1983 and Dyer 1983). To perform plan recognition, the reader must, among other things, realize when character's intentions are being communicated; this may involve realizing that a particular sentence is subjective rather than objective. In (7), a represented thought communicates a character's intentions:

> (7)
> $^{7.1}$He [Jeff] could see her walking the other way. $^{7.2}$If he wanted to avoid notice, he would have to act with the same deliberate manner as all the robots around him. $^{7.3}$He lengthened his stride and gave chase without otherwise altering his body language. [Wu, *Cyborg*, p. 71]

Sentence (7.2) is not a purely narrative sentence informing the reader that if Jeff wants to avoid notice, he must change the way he is walking. Rather, (7.2) is Jeff's thought expressing his goal to avoid notice, and also his belief about how to achieve it. Understanding this is necessary for the reader to perceive the intentionality behind the action described in (7.3).

In addition to being important for plan recognition, tracking POV is necessary to distinguish what is true in the fictional world from what is believed by the characters. While objective sentences are unquestionably true in the fictional world, subjective sentences reflect the subjective character's beliefs, which may be false in the fictional world (Uspensky 1973, Kuroda 1976, Cohn 1978, Banfield 1982). A striking example is the following:

> (8) This was David's boy. [Bridgers, *All Together Now*, p. 91]

Sentence (8) is actually about a female character named "Casey". However, it is the represented thought of a character who believes that Casey is a boy, and reflects this false belief about her.

Beliefs and intentions are just two of many types of attitudes communicated by subjective sentences (see Dyer 1983 for a processing model of how various attitudes can be related to goals). The algorithm for tracking POV presented in this paper is intended to be just one part of an overall narrative-understanding system. Identifying which attitudes are being presented, and incorporating them into a representation of the meaning of the text, are beyond the scope of this work. But in order to accomplish such tasks, the system must recognize *when* attitudes are being presented in the first place and identify the character whose attitudes they are; these are the tasks involved in tracking the psychological POV.

POV also has important implications for discourse processing. Discussion of this topic is deferred until section 14.

Sentences that present[1] the attitudes of someone who is not the writer appear in genres other than third-person fictional narrative text, such as newspaper articles and text-books (Banfield 1982), even if we might prefer another term than "represented thought" for such sentences. Thus, the problem of tracking POV arises for these other genres as well. Following is a passage from *The New York Times* (1990) that contains a "represented thought":

> Looking at the more severely affected countries, experts are wondering where the saturation point will be. Where will the infection rate level off as most of those engaging in riskier behavior fall prey: 30 percent? 40 percent? [*The New York Times*, September 16, 1990]

The second sentence presents something that the experts are wondering; 30 and 40 percent are the experts' guesses, not those of the writer.

Third-person fictional narrative text is the focus of this work primarily because it contains so many prototypical instances of sentences that present attitudes of someone who is not the writer. Further, by restricting the types of texts considered (see section 2.2 above), it is possible to constrain the problem to choosing among an objective POV and the points of view of the characters. In genres such as soft news and editorials, for example, the writer may present his or her own attitudes; when attitudes of the writer, the speaker, or a narrator may be presented, the problem is less constrained. Hamburger (1973), Kuroda (1973), Banfield (1982), Galbraith (1990), and Wiebe (1991) address this issue more fully. Finally, there is a great deal of previous work to build upon, in linguistics and especially literary theory, investigating POV in novels and short stories.

## 4. Approach

*Reasoning* about whether a sentence is objective or the subjective sentence of this or that character is certainly part of tracking POV (Fillmore 1974). But it is reasonable to hypothesize that, in the face of all of the inferential possibilities, discourse expectations too are involved in tracking POV; that is, in the absence of an explicit indication of POV, readers are intended to assume that POV is being manipulated in one of the usual ways, and to try to interpret the sentence accordingly. (Similar suggestions have been made by Carberry (1989) with respect to resolving intersentential ellipsis and by Sidner (1983) with respect to pronoun resolution.) We can view the text as composed of maximal blocks of objective sentences (**objective contexts**) and maximal blocks of subjective sentences that have the same subjective character (**subjective contexts**). Further, we can view the process of tracking POV as recognizing the following discourse operations:

1. Sentence $s$ **continues** the current POV: $s$ and the previous sentence are either both objective or both subjective sentences of the same character.

2. Sentence $s$ **resumes** $y$'s POV: $s$ is a subjective sentence of $y$ and is preceded by an objective context, which is in turn preceded by a subjective context of the same character $y$.

3. Sentence $s$ **initiates** $y$'s POV: $s$ is a subjective sentence of $y$ and either $s$ is the first subjective sentence of a scene, or the SC of the previous subjective sentence is a different character $z$.

The approach taken in this work is to seek, by extensive examination of naturally occurring narratives, regularities in the ways that authors initiate, resume, and continue a character's point of view, and to develop an algorithm that tracks point of view on the basis of the regularities found. Given certain combinations of sentence features (e.g., tense, aspect, lexical items that potentially express subjectivity, the types of states of affairs that the sentence is about, and the identities of the actors or experiencers of those states of affairs), and of the current context (e.g., whether the previous sentence was subjective or objective, whether a paragraph break separates the current and previous sentences, and the identity of the SC of the previous subjective sentence, if there was one), particular point-of-view operations can be expected. A simple example: a sentence that (1) is about an action, (2) is in the past progressive, and (3) follows, without a paragraph break, the subjective sentence of a character $c$, usually continues $c$'s POV.

The examination of texts mentioned above was not a formal empirical investigation, so the algorithm should be viewed as a hypothesis that can be subjected to such tests. However, I examined passages from over forty novels and short stories to identify the regularities upon which the algorithm is based, and strictly adhered to the practice of considering only naturally occurring examples in developing the algorithm. Further, some preliminary empirical tests of the algorithm have been performed (see section 12), among them psychological experiments of specific aspects of the algorithm. The results of these experiments are positive. I describe them as "preliminary" so as not to suggest that the entire algorithm has already been subjected to psychological experimentation.

## 5. Previous Work

There are literary theorists and linguists who investigate linguistic aspects of subjective sentences. The present work greatly benefited from their investigations, most directly from Doležel (1973), Uspensky (1973), Kuroda (1973 and 1976), Fillmore (1974), Cohn (1978), and especially Banfield (1982). However, the relevant work in the above fields is descriptive only; it describes characteristics of subjective sentences, but does not address the problem of *tracking* POV. An exception is work on POV and aspect that shows that aspect is only a context-sensitive marker of subjectivity (Ehrlich 1987 and 1990, and Caenepeel 1989; see section 9).

In AI, Nakhimovsky (1988) suggests a discourse-processing approach to tracking POV, but does not develop it in any depth. Also, Reiser (1981) simply suggests that POV may be established by syntactic clues, and by including "more episodes and internal information" about a character (p. 209).

A great deal of work in AI has involved inferring speaker and hearer attitudes in conversation (such as work in plan recognition in conversation, see, e.g., Allen and Perrault 1980; in user modeling, see, e.g., the papers in the special issue of *Computational Linguistics* on user modeling 1988; and in dynamically constructing nested-belief environments during understanding, see, e.g., Wilks and Bien 1983) and character's attitudes in stories (see, e.g., Wilensky 1983 and Dyer 1983). Such beliefs might be *about* what agents who are mentioned in the discourse believe, but the question is not addressed as to whether an utterance itself presents what is actually the object of an agent's attitude, where that agent is not mentioned in the sentence.

In summary, there is no previous detailed investigation of the problem of tracking the psychological POV.

## 6. Overview of the Algorithm

The algorithm will be developed in detail in subsequent sections. This section provides an overview, giving the input and output of the basic components. While this involves using some terms before they are defined, it will provide the reader with a framework in which to understand the material that follows. As well, it will enable me to clarify the

```
i ← 1
context ← ⟨{}, {}, {}, presubjective-nonactive⟩
loop
    if ¬SENTENCE(ITEM(text, i)) then
        context ← NEW-CONTEXT'(ITEM(text, i), context)
    else
        interpretation ← POV(FEATURES(ITEM(text,i)), context)
        context ← NEW-CONTEXT(interpretation, context)
    end if
    i ← i + 1
end loop
```

**Figure 1**
Overview.

focus of the work and to state exactly what has been implemented.

Figure 1 gives the algorithm at the highest level and a corresponding flow-diagram. Of the functions shown in figure 1, it is functions POV, NEW-CONTEXT, and NEW-CONTEXT' that are addressed in this work. Functions ITEM and FEATURES, preprocessing functions, represent a subset of the other tasks performed by an overall NLU system. The remainder of this section gives the input/output mappings of all five functions.

### 6.1 The Preprocessing Functions

• Function ITEM maps the text and the current position in the text into the *input item* at that position, i.e., the paragraph break, scene break (see section 8.2.1), or sentence (see section 6.3.2) that is at the current position:

$$\text{ITEM} : Text \times Position \rightarrow InputItem.$$

This function is not implemented in the system, so a facility is provided enabling the user to input the current input item.

• Function FEATURES maps a sentence (see section 6.3.2 below) into a set of features:

$$\text{FEATURES} : Sentence \rightarrow FeatureSet.$$

A *FeatureSet* consists of the following (some of the features will be expanded upon in later sections, as indicated; for extensive detail, see Wiebe 1990):

1. The **potential subjective elements** in the sentence, if any. The potential subjective elements form a large class that includes lexical items used with particular meanings, tense, aspect, and certain syntactic properties (see section 9.2).

2. The type of state of affairs that each clause is about. For example, the main clause of the following is about a private state and the subordinated clause is about an action:

    John wondered whether Mary opened the door.

    Note that each of the following is about a private state:

    Mary was afraid of the dark.
    The darkness made Mary afraid.

    Wiebe 1990 gives a list of private-state terms, noting which syntactic roles the experiencer fills in various clause structures (this material is drawn from Quirk et al. 1985). Section 6.3.1 gives the categorization of states of affairs used in this work.

3. An indication of whether or not the head noun of the subject of the main clause is a private-state noun (e.g., 'pain' and 'astonishment') and, if it is, the state of affairs that that noun is about. Examples of sentences with such private-state nouns are:

    (10) The feeling went away.
    (11) The pain increased.
    (12) His astonishment grew.

4. The experiencers and actors (i.e., particular case fillers (Fillmore 1968)) of the states of affairs of items (2) and (3). Because a single action might have more than one actor, and a single state might have more than one experiencer, each of these is a set. If an experiencer or actor is not mentioned in the sentence (as is the case for the subject nouns in (10) and (11)), then that experiencer or actor is the empty set.

5. An indication of whether or not the sentence contains a narrative parenthetical, and, if it does, the identity (or identities) of the individual(s) referred to by the subject of the parenthetical.

6. Some other syntactic information not included above.

Function FEATURES involves the resolution of syntactic, semantic, and discourse/pragmatic ambiguities that are outside the scope of this work (see section 14 for a discussion of interactions between point of view and discourse processing). While the implemented system demonstrated in appendix II takes actual sentences as input, it does not truly implement FEATURES, but is successful in computing the *FeatureSet* of a sentence only for sentences that fall within its limited coverage. There is another version of the system that queries the user for the information returned by function FEATURES, to enable the algorithm to be tested on unlimited text, without concern for problems not addressed in this work. It is this system that is used in the test of the algorithm presented in appendix III and to support the psychological experiments mentioned in sections 8.3.1, 9.3, and 12.

## 6.2 The Central Functions

• Function POV, which is implemented, maps a *FeatureSet* and a *Context* into an *Interpretation*:

$$\text{POV} : FeatureSet \times Context \rightarrow Interpretation.$$

A *Context* and an *Interpretation* are as follows. A *Context* consists of (1) the identity of the SC of the last subjective sentence that appeared in the text, if there was one, (2) the identity of the last **active character** (defined in section 8.2.2), if there was one, (3) the identities of any characters whose points of view were taken earlier in the text, and (4) the current *text situation* (defined just below in section 6.3.4):

$$\begin{aligned}Context \quad = \quad &\langle LastSC, \\ &LastActiveCharacter, \\ &PreviousSCs, \\ &TextSituation\rangle\end{aligned}$$

An *Interpretation* is either that the sentence is the subjective sentence of a particular character, or that the sentence is objective and has a particular active character (*ActiveCharacter* is the empty set for objective sentences without active characters):

$$\begin{aligned}Interpretation \quad \in \quad &\{\langle \text{objective}, ActiveCharacter\rangle \mid ActiveCharacter \subseteq Characters\} \cup \\ &\{\langle \text{subjective}, SC\rangle \mid SC \subseteq Characters\}.\end{aligned}$$

• Functions NEW-CONTEXT and NEW-CONTEXT′ return the *Context* of the next input item, the former for the case in which the current item is a sentence, the latter for the case in which the current item is a scene or paragraph break:

$$\text{NEW-CONTEXT} : Interpretation \times Context \rightarrow Context.$$

$$\text{NEW-CONTEXT}' : \{\text{paragraph break, scene break}\} \times Context \rightarrow Context.$$

These functions are also implemented; algorithms for them follow trivially from the definitions of an interpretation and of a context and its components.

The *Context* of the $i^{th}$ input item in text $t$, $c_i$, is

$$c_i = \begin{cases} \langle\{\},\{\},\{\},\text{presubjective-nonactive}\rangle & \text{if } i = 1 \\[1ex] \text{NEW-CONTEXT}'(\text{ITEM}(t, i-1), c_{i-1}) & \text{if } i > 1 \ \& \\ & \neg\text{SENTENCE (ITEM } (t, i-1)) \\[1ex] \text{NEW-CONTEXT}( & \text{if } i > 1 \ \& \\ \quad \text{POV}(\text{FEATURES}(\text{ITEM}(t, i-1)), c_{i-1}), & \text{SENTENCE (ITEM } (t, i-1)) \\ \quad c_{i-1}) & \end{cases}$$

where "presubjective-nonactive" is the text situation of the first item in a text, and SENTENCE($x$) is true iff $x$ is a sentence. To refer to the components of a context, we shall use the functions LAST-SC-OF, LAST-ACTIVE-CHARACTER-OF, PREVIOUS-SCS-OF, and TEXT-SITUATION-OF, which map a context $C$ to the *LastSC*, *LastActiveCharacter*, *PreviousSCs*, and *TextSituation* of $C$, respectively.

In the above definitions, *LastSC*, *LastActiveCharacter*, *SC*, and *ActiveCharacter* are sets, and *PreviousSCs* is a set of sets. This is because a subjective sentence can represent the shared psychological POV of more than one physical character (Banfield 1982). For example:

(13)
    [13.1]Leaning out of the window side by side the two women watched the man ... [13.2]Now he threw away his cigarette. [13.3]They watched him. [13.4]What would he do next? [Woolf, *The Years*, p. 103; cited by Banfield 1982, p. 96]

Sentence (13.4) is a represented thought whose SC is more than one physical character.

**6.3 States of Affairs, Sentences, and Contexts: some further details**

**6.3.1 States of Affairs.** The following are the range of state-of-affairs types that can be included in a *FeatureSet*. For more refined categorizations for the purpose of analyzing tense and aspect, see, for example, Reichenbach 1947 and the papers in the special issue of *Computational Linguistics* on tense and aspect 1988.

1. private-state actions, such as looking and sighing (see section 11);
2. other kinds of actions;
3. private states; and
4. nonprivate states, such as being six feet tall.

The rough part of this categorization is the boundary between the second and fourth items. We shall assume that function FEATURES classifies only clear instances of nonprivate states as such, and that states of affairs that fall between the categories of action and nonprivate state, such as processes, are classified as actions.

    In quoted speech, there are states of affairs on two levels: those spoken of and the action of speaking itself. The regularities on which the algorithm is based involve the latter. Thus, the contents of the quoted string are not considered, and, even if a **discourse parenthetical** does not actually appear, the sentence is viewed as one whose main clause is about a communicative action (a subtype of (2) above), that is, as one whose main verb phrase contains a communicative verb with a quoted string as object. We return to this point in section 9.5.

**6.3.2 Sentences.** We shall call any *InputItem* that is not a scene or paragraph break a "sentence", although such an *InputItem* is sometimes smaller than an actual sentence, sometimes larger. The former occurs for compound sentences and the latter occurs for sentences of quoted speech.

    Suppose that a compound sentence with conjuncts $c_1, \ldots, c_n$ starts at position $i$ in text $t$. Then, ITEM($t,i$) = $c_1$, ..., ITEM($t,i+n\text{-}1$) = $c_n$. We shall call each $c_i$ a "sentence".

    In the case of quoted speech, everything enclosed within a single pair of quotes, together with the accompanying discourse parenthetical, if there is one, counts as a single *InputItem* which we shall call a "sentence". For example, (4.4) and (4.5) together compose a single *InputItem* (a single "sentence"). Since the algorithm does not consider the contents of quoted strings (see section 6.3.1 above), there is no reason for each constituent sentence of a quoted string to be a separate *InputItem*.

    From this point forward, numbering within cited passages is as follows: each *InputItem* that is not a paragraph or scene break is given a separate number.

    Now that the units of input have been specified, the following can be noted. In this work, if any part of a sentential *InputItem* $s$ is subjective, then $s$ as a whole is considered to be subjective. This makes the algorithm easier to understand. Enabling it to report which part of $s$ is subjective, if relevant, would involve straightforward refinements of the

text situations, of interpretations, and of some conditional steps of the algorithm into subcases.

**6.3.3 Choosing a State of Affairs.** Out of the states of affairs included in a *FeatureSet* for a sentential input item $s$, the algorithm chooses just one to consider. Specifically, the algorithm for POV uses the following function:

$$\text{CHOSEN-STATE-OF-AFFAIRS} : FeatureSet \rightarrow StateOfAffairs.$$

As we shall see in section 8, private states are particularly important to consider, because if a sentence that is about a private state is interpreted to be a private-state report, then the SC of the sentence is the experiencer of the private state. A subordinated clause can report a character's private state; an example is "thinking ..." in the following sentence (as it appears in the novel):

> When he [Call] got within fifteen miles of Lonesome Dove he cut west, thinking they would be holding the herd in that direction. [McMurtry, *Lonesome Dove*, p. 181]

Following is the specification of CHOSEN-STATE-OF-AFFAIRS. Let $c_{main}$ be the main clause of $s$; let $c_1, c_2, \ldots, c_n$ be the other $n$ clauses of $s$; and let $hn$ be the head noun of the subject of $c_{main}$. Further, let $soa_{main}$ be the state of affairs that $c_{main}$ is about; let $soa_{hn}$ be the state of affairs that $hn$ is about (but if $hn$ is not about a state of affairs—for example, if it is a proper noun—then let $soa_{hn}$ be nil); and let $soa_i$ ($1 \leq i \leq n$) be the state of affairs that $c_i$ is about. Then, the result of CHOSEN-STATE-OF-AFFAIRS(FEATURES($s$)) is as follows:

**If** $soa_{main}$ is a private state **then**
    the result is $soa_{main}$
**else if** $soa_{hn}$ is a private state **then**
    the result is $soa_{hn}$
**else if** $\exists\ soa_i$ ($1 \leq i \leq n$) such that $soa_i$ is a private state, and $c_i$ is not
        subordinated to another clause $c_k$ such that $soa_k$ is a private state **then**
    the result is $soa_i$ (if there is more than one such $soa_i$, one is randomly chosen)
**else** the result is $soa_{main}$.

Table 1 gives some examples.
    Note that private-state terms appearing in certain types of constituents cannot be used to report private states. An example is a **manner adverbial** (Quirk et al. 1985), such as the italicized portion of "Japheth turned the book over *in a puzzled manner*." (There are others addressed in Wiebe 1990, but due to space limitations, we shall ignore them in this paper.) The phrase "in a puzzled manner" does not *report* a private state, but rather describes the manner in which something is done. The state of affairs chosen for this sentence is simply the main-clause action.
    To facilitate discussion, a sentential *InputItem* of which the state of affairs chosen for consideration is of type $X$ will be called an "$X$ sentence", for example, "action sentence" and "private-state sentence".

**6.3.4 The Text Situations.** Following are the text situations (recall that a text situation is part of the context, defined in section 6.2).[2] To make it easier to understand them, definitions are given in both English and diagrams. Each diagram shows what appears between the start of the current scene, represented by "start-of-scene", and the current position, represented by a diamond ($\diamond$). The start of the current scene is either a scene break or the very beginning of the text. The symbol "¶" represents a paragraph break; "objective-sentence" and "subjective-sentence" represent objective and subjective sentences, respectively; "sentence" alone represents either a subjective or objective sentence; the symbol "*" means 0 or more occurrences, "+" means 1 or more occurrences, and "..." represents any number of paragraph breaks and sentences (but not scene breaks, since only what has appeared since the start of the current scene is shown). A scene is assumed to always begin with a paragraph break.

| | |
|---|---|
| $s$: | "Japheth turned the book over in a puzzled manner." |
| Chosen: | $soa_{main}$ |
| $s$: | "The pain increased." |
| Chosen: | $soa_{hn}$ |
| $s$: | "The pain angered him." |
| Chosen: | $soa_{main}$ |
| $s$: | "When he got within fifteen miles of Lonesome Dove he cut west, thinking they would be holding the herd in that direction." |
| Chosen: | $soa_i$, where $c_i$ = "thinking they would be holding the herd in that direction" |
| $s$: | " 'Rosie's just saying that. She doesn't really care,' Zoe said." (sentence (15.3)). |
| Chosen: | $soa_{main}$, which is the action of Zoe saying the quoted string. |
| $s$: | "What are you doing in here?" (sentence (4.1)). |
| Chosen: | $soa_{main}$, which, as $s$ appears in the novel, is Zoe saying the quoted string. That Zoe is the speaker is in FEATURES($s$) (see items (2) and (4)) of section 6.1 and the end of section 6.3.1). |

**Table 1**
Choosing a state of affairs: examples.

1. **presubjective-nonactive**: a subjective sentence has not appeared so far in the current scene, and a sentence with an **active character** (defined in section 8.2.2 below) has not appeared so far in the current paragraph.
   start-of-scene  ¶  (objective-sentence$^+$ ¶)*  $\underbrace{\text{objective-sentence}^*}_{\substack{\text{None has an} \\ \text{active character}}}$  $\diamond$

2. **presubjective-active**: a subjective sentence has not appeared so far in the current scene, but a sentence with an active character *has* appeared earlier in the current paragraph.
   start-of-scene  ¶  (objective-sentence$^+$ ¶)*  $\underbrace{\text{objective-sentence}^+}_{\substack{\text{At least one has} \\ \text{an active character}}}$  $\diamond$

3. **continuing-subjective**: the current sentence follows a subjective sentence without a paragraph break.
   start-of-scene  ...  ¶  sentence*  subjective-sentence  $\diamond$

4. **broken-subjective**: the current sentence follows a subjective sentence, but *after* a paragraph break.
   start-of-scene  ...  subjective-sentence  ¶  $\diamond$

5. **interrupted-subjective**: the current sentence follows an objective sentence, but an earlier sentence in the current paragraph is subjective.
   start-of-scene ... ¶ sentence* subjective-sentence sentence* objective-sentence ◇

6. **postsubjective-nonactive**: a subjective sentence has appeared in the current scene, and an objective sentence and a paragraph break have appeared since the last subjective sentence. However, a sentence with an active character does not appear earlier in the current paragraph.
   start-of-scene ... subjective-sentence ... objective-sentence ¶ ◇

   -or-

   start-of-scene ... subjective-sentence ... ¶ $\underbrace{\text{objective-sentence}^+}_{\substack{\text{None has an} \\ \text{active character}}}$ ◇

7. **postsubjective-active**: like the postsubjective-nonactive situation, except that a sentence with an active character does appear earlier in the current paragraph.
   start-of-scene ... subjective-sentence ... ¶ $\underbrace{\text{objective-sentence}^+}_{\substack{\text{At least one has} \\ \text{an active character}}}$ ◇

## 7. Focus of this Work

Algorithms for the preprocessing functions are not given in this paper, and algorithms for NEW-CONTEXT and NEW-CONTEXT$'$ are given in appendix I. The remainder of this paper develops an algorithm for function POV, which at the highest level is the following:

POV (*featureSet, context*)
**if** SENTENCE-IS-SUBJECTIVE (*featureSet, context*) **then**     §9
    **return** ⟨subjective, IDENTIFY-SC (*featureSet, context*)⟩     §8
**else**
    **return** ⟨objective, ACTIVE-CHARACTER-OF (*featureSet, context*)⟩     §8
**end if**

Active characters are discussed in the section about identifying the SC (section 8), because the raison d'etre of the active-character component of an interpretation is that the active character of an objective sentence may become the *LastActiveCharacter* of a later context, $context_i$, and as such, may become the SC of a subjective sentence that is *processed in $context_i$*.

## 8. Identifying the Subjective Character

### 8.1 Introduction
The SC of a subjective sentence can always be identified from the sentence itself if the sentence has a narrative parenthetical, such as 'Dennys thought' in (2), and can sometimes be so identified if the sentence is about a private state. When the SC is not identifiable from the sentence, she is often a previously mentioned character. Thus, as the text is processed, any algorithm for tracking POV must keep track of characters who are likely to become the SC of a later subjective sentence. I call such characters **expected subjective characters**. The algorithm presented in this paper considers two possibilities: the last subjective character and the last active character of the current context. Each is an expected subjective character only in certain text situations. The idea of keeping track of entities evoked in the text in order to interpret later sentences is due to work on anaphora resolution (e.g., Sidner 1983 and Grosz, Joshi, and Weinstein 1983). Active characters are based specifically on Sidner's actor focus, but while Sidner's actor focus is "whoever is currently the agent in the sentence" (p. 282), many sentences with agents do not have active characters, as we shall see below in section 8.2.2.

In some cases, the SC of a subjective sentence is not identifiable when the sentence appears, but may be identifiable after later sentences are processed. The algorithm, as presented in Wiebe 1990, handles one such case. Due to space limitations, however, this aspect of the algorithm is not presented in this paper, but is only briefly described (in section 8.2.4).

Here is a high-level algorithm for identifying the SC (comments are preceded by '%'):

IDENTIFY-SC(*featureSet*, *context*)
% IDENTIFY-SC(*featureSet*, *context*) ⊆ *Characters*; the empty set indicates failure.
**if** IDENTIFY-SC-FROM-THE-SENTENCE(*featureSet*, *context*) ≠ {} **then**
    **return** IDENTIFY-SC-FROM-THE-SENTENCE(*featureSet*, *context*)
**else if** LAST-SC-IS-AN-EXPECTED-SC (*context*) **and**
    LAST-ACTIVE-CHARACTER-IS-AN-EXPECTED-SC(*context*) **then**
    **return** CHOOSE-AN-EXPECTED-SC(*featureSet*, *context*)
**else if** LAST-SC-IS-AN-EXPECTED-SC(*context*) **then**
    **return** LAST-SC-OF(*context*)
**else if** LAST-ACTIVE-CHARACTER-IS-AN-EXPECTED-SC(*context*) **then**
    **return** LAST-ACTIVE-CHARACTER-OF(*context*)
**else**
    **return** {}
**end if**

The following subsections refine and illustrate the above algorithm. We first consider identifying the SC based on expected subjective characters (in section 8.2), and then identifying the SC from the sentence (in section 8.3).

### 8.2 Identifying the SC from the Context

In this section, we consider cases in which the SC of a subjective sentence cannot be identified from the sentence itself. We first consider the last subjective character (section 8.2.1), then the last active character (section 8.2.2 ), and then the cases in which both (section 8.2.3) or neither (section 8.2.4) are expected subjective characters.

**8.2.1 The Last Subjective Character.** An SC who is not identifiable from the sentence itself is most often the last subjective character. In this case, the current sentence is continuing a character's POV, if the previous sentence was also subjective, or resuming one, if objective sentences have appeared since the last subjective sentence. Sentence (1.2) above illustrates the former and passage (15) illustrates the latter:

> (15)
> $^{15.1}$Zoe would have liked to punch her. $^{15.2}$She could not understand why her parents didn't know Rosie was a phony.
>     $^{15.3}$"Rosie's just saying that. She doesn't really care," Zoe said.
>     $^{15.4}$"I do too!" cried Rosie.
>     $^{15.5}$"Phony!" Zoe yelled.

$^{15.6}$"That will be enough." $^{15.7}$Their father stood up. $^{15.8}$"You may take your plate to the kitchen."

$^{15.9}$"What about Rosie!" Zoe yelled.

$^{15.10}$"*I* will worry about Rosie."

$^{15.11}$There was no use arguing. [Oneal, *War Work*, p. 40; italics in original]

Sentences (15.1), (15.2), and (15.11) are Zoe's subjective sentences. Sentence (15.11) expresses Zoe's judgment that there is no use in arguing, resuming Zoe's point of view: it has the same SC as the last subjective sentence, (15.2), and is separated from (15.2) by objective sentences (15.3)-(15.10).

If there has not been a subjective sentence so far in the text, then the last subjective character, which is the empty set in this case, clearly should not be an expected subjective character. Moreover, drastic spatial and temporal discontinuities can block the continuation or resumption of a character's psychological POV. This paper considers one such kind of discontinuity, a **scene break**. The condition under which the last subjective character of $context_i$ is an expected subjective character is when a subjective sentence has appeared in the current scene. That is, LAST-SC-IS-AN-EXPECTED-SC ($context_i$) is true iff TEXT-SITUATION-OF($context_i$) $\notin$ {presubjective-nonactive, presubjective-active}.

A scene break is a break from one parallel story-line to another. Almeida (1987) analyzes parallel story-lines as forming separate **narrative-lines**, which are stretches of narrative that are controlled by single **now-points**. The following passage illustrates the situation in which a scene break blocks the resumption of a character's point of view:

(16)
$^{16.1}$Moving fast, in the dark.

$^{16.2}$He'd lost Cherry. $^{16.3}$He'd lost the hammer. $^{16.4}$She must've slid back down into Factory when the guy fired his first shot. $^{16.5}$Last shot, if he'd been under that box when it came down …

[Mixture of subjective sentences of the same character and quoted speech]
[Chapter break]

$^{16.6}$What kind of place was this, anyway?

$^{16.7}$Things had gotten to a point where Mona couldn't get any comfort out of imagining Lanette's advice. [Gibson, *Mona Lisa Overdrive*, pp. 275-276; ellipsis in original]

As passage (16) appears in the novel, the SC of all of the subjective sentences before the chapter break is the character Slick, and a scene break occurs at the chapter break. While the sentence following the break is subjective, the SC of that sentence should *not* be identified to be the last subjective character, Slick.

**8.2.2 The Last Active Character.** A SC who is not identifiable from the sentence itself may also be the actor of an action that a previous objective sentence is about (but less commonly than the last subjective character). Since this character need not be the SC of the last subjective sentence, this is a way for the author to initiate a new point of view. Following is an example (by this point in the novel, both Jake and Augustus have been the SC of previous subjective sentences):

(17)

$^{17.1}$Jake felt sour. $^{17.2}$He wished again that circumstances hadn't prompted him to come back. $^{17.3}$He had already spent one full night on horseback, $^{17.4}$and now the boys were expecting him to spend another, all on account of a bunch of livestock he had no interest in in the first place.

$^{17.5}$"I don't know as I'm coming," he said. $^{17.6}$"I just got here. If I'd known you boys did nothing but chase horses around all night, I don't know that I would have come."

$^{17.7}$"Why, Jake, you lazy bean," Augustus said, $^{17.8}$and walked off. $^{17.9}$Jake had a stubborn streak in him, $^{17.10}$and once it was activated even Call could seldom do much with him. [McMurtry, *Lonesome Dove*, p. 162]

As this passage appears in the novel, (17.1)-(17.4) are the subjective sentences of Jake, and (17.9)-(17.10) are the subjective sentences of Augustus, the actor of an action that a previous objective sentence was about (sentence (17.8)). However, it is Jake who is the last subjective character, so Augustus's point of view is being initiated, not merely resumed or continued.

The situation I observed in which POV shifts to an actor (who is not also the last subjective character) is one in which the actor was the SC of some previous subjective sentence in the text, and the sentence about his or her action is focused by the text. The precise situation is captured by the following specifications of what an active character is, what the last active character is, and of the text situations in which the last active character is expected.

Suppose that $os$ is an objective sentence that is $InputItem_j$. Saying that $os$ has an active character means that POV(FEATURES($os$), $context_j$) = ⟨objective, $ac$⟩, where $ac$ is *not* the empty set. This is the case iff:

i. $os$ is about an action that is actually performed in the current scene by $ac$ (more precisely, when the state of affairs chosen for consideration is such an action), and

ii. $ac \subseteq$ PREVIOUS-SCS-OF ($context_j$). That is, $ac$ is in the set of characters who have been the SC of some previous subjective sentence in the text (possibly before the current scene).

The algorithm for POV determines whether an action meets the conditions in (i) by looking at such things as the tense, aspect, and mood of $os$ (the features the algorithm considers are in FEATURES($os$)). First, to guarantee that the action is not performed earlier or later than the current moment in the story, the main verb phrase of $os$ must be in the simple past. Also, to be about a specific action, $os$ cannot be **habitual**. So, the main verb phrase cannot be accompanied by an adverbial such as *at times, usually, rarely*, or *on weekends*.³ Finally, to be about an action that actually occurs, the main clause of the sentence must not contain modal auxiliary verbs such as *could, going to, had better, have to, might, should,* or *must,* modal adverbs such as *likely, maybe, perhaps,* or *possibly*, and it must not be negated. If the action is quoted speech, then these restrictions apply to the discourse parenthetical.

Now we turn to the *last* active character. In $context_i$, LAST-ACTIVE-CHARACTER-OF($context_i$) is the empty set if no sentences with active characters have appeared; otherwise, it is the active character of the last sentence that had one.

The last active character is an expected subjective character only when a subjective sentence has not appeared earlier in the current paragraph, and there is an earlier sentence in the current paragraph that has an active character. That is, LAST-ACTIVE-CHARACTER-IS-AN-EXPECTED-SC ($context_i$) is true iff TEXT-SITUATION-OF($context_i$) ∈ {presubjective-active, postsubjective-active}.

**8.2.3 When There Are Two Expected Subjective Characters.** When the last active character and the last subjective character are both expected subjective characters (which, as the reader may have noticed, is when the current text situation is postsubjective-active), the algorithm chooses the last active character in most cases, since he or she is more highly focused by the text. In fact, there is only one case in which the algorithm chooses the last *subjective* character: when the sentence is about the last active character (specifically, when the last active character is the experiencer or actor of the state of affairs chosen for consideration). Following is the algorithm for function CHOOSE-AN-EXPECTED-SC introduced in section 8.1 above. The only new function is EXPERIENCER-OR-ACTOR-OF, which maps a state of affairs *soa* and a feature set $f$ into the actor or experiencer of *soa* (if *soa* or its actor or experiencer is not in $f$, the result is the empty set).

CHOOSE-AN-EXPECTED-SC(*featureSet*, *context*)
**if** EXPERIENCER-OR-ACTOR-OF (CHOSEN-STATE-OF-AFFAIRS (*featureSet*), *featureSet*)
    = LAST-ACTIVE-CHARACTER-OF (*context*) **then**
    **return** LAST-SC-OF (*context*)
**else**
    **return** LAST-ACTIVE-CHARACTER-OF (*context*)
**end if**

The criterion for choosing the last subjective character is correct for the situation in which the last *subjective* character's attention is directed toward the last *active* character, and the sentence represents the last *subjective* character's reflection about or observation of the other. It is incorrect, however, if the sentence is the last *active* character's self-reflection or self-perception; this heuristic relies on the relative infrequency of subjective sentences about oneself.

Consider subjective sentence (15.11). When this sentence is encountered, only the last subjective character (Zoe) is expected, because the sentence is at the beginning of a new paragraph. The algorithm correctly identifies her to be the subjective character. Now consider (17.9), which is also subjective. When it is encountered, not only is the last subjective character expected (Jake), but so is the last active character (Augustus): Augustus is the active character of (17.7)-(17.8), because he has been the SC of previous subjective sentences and (17.7)-(17.8) are objective sentences about his current actions; and, when (17.9) is encountered, the last active character is expected, since (17.8) (but no subjective sentences) appeared earlier in the current paragraph. The algorithm correctly identifies the SC to be Augustus (the last active character), rather than Jake (the last subjective character), because the criterion for choosing the last *subjective* character is not satisfied: the sentence is not about the last active character. Competition also arises in the following passage, but this time it is the last subjective character who should be chosen. When the passage is encountered, Lorena is the last subjective character:

> (18)
> $^{18.1}$"I never tolt on you, Lorie," he [Lippy] said. $^{18.2}$He looked like he might cry too. $^{18.3}$You'll just have to cry, she [Lorena] thought. [McMurtry, *Lonesome Dove*, p. 218]

By this point in the novel, Lippy has been a subjective character. Thus, since (18.1) is about his current action, he is the active character of (18.1). After (18.1), Lippy, as the last active character, is expected, because (18.1) is an objective sentence that begins a new paragraph. Sentence (18.2) is subjective because the evidential 'looked like' appears. Competition is (correctly) resolved in favor of the last subjective character (Lorena), because the sentence is about the last active character (Lippy).

**8.2.4 When There Are No Expected Subjective Characters.** If no character is expected, then the algorithm fails to identify the SC at this point in the text. This eventuality is rare, relative to others: it can arise only upon the first subjective sentence of a scene (otherwise, the last subjective character would be expected) and only in the absence of one of the things that are usually used to initiate a character's point of view (such as a narrative parenthetical or private-state report, discussed below in section 8.3). An example of this is (16.6), and another is (19.1), which is the beginning of a novel:

> (19)
> $^{19.1}$Captain Scalawag's treasure! $^{19.2}$It was the first thing Pete thought of when he woke up. [Lorimer, *The Mystery of the Missing Treasure*, p. 1]

In a case such as (19.1), it is not possible to identify the SC without reading further in the text. In a case such as (16.6), however, it might be possible to do so. That is, a reader might be able to infer who the SC is from clues in the sentence, such as indications of place, or facts that only a certain character knows. This process is not addressed in this work. However, the author could have made identifying the SC easier by using, for example, a narrative parenthetical or private-state report (see section 8.3); by not including one of these, the author is deliberately demanding some extra work from the reader.

The algorithm as presented in Wiebe 1990 can identify the SC after later sentences are processed in the case where a later sentence contains a narrative parenthetical or is a private-state report. As illustrations, (16.7) is a private-state report whose SC, Mona, is the SC of (16.6), and (19.2) is a private-state report whose SC, Pete, is the SC of (19.1).

**8.3 Identifying the SC from the Sentence**
We now turn to cases in which the SC is identifiable from the sentence itself (that is, cases in which IDENTIFY-SC-FROM-THE-SENTENCE(*featureSet, context*) $\neq \{\}$; see section 8.1 above). In these cases, the SC of sentence $s$ is chosen from among certain characters in FEATURES($s$). Such a character need not be the last subjective character; when she is not, $s$ initiates her POV. Thus, the cases discussed in this section—i.e., uses of sentences with certain features in particular contexts—are ways to initiate a character's POV.

The straightforward case is when $s$ contains a narrative parenthetical, such as sentence (2). The SC is always the subject of the parenthetical.

The less straightforward case is when $s$ is a private-state sentence. Doležel (1973), Cohn (1978), and Banfield (1982) all note that a private-state sentence is a way to initiate a character's POV. In the framework presented in this paper, the SC may be the experiencer of the private state, even if she is not the last subjective character. An example occurs in (20):

> (20)
> $^{20.1}$"Drown me?" Augustus said. $^{20.2}$"Why if anybody had tried it, those girls would have clawed them to shreds." $^{20.3}$He knew Call was mad, [20.4]but wasn't much inclined to humor him. [20.5]It was his dinner table as much as Call's, [20.6]and if Call didn't like the conversation he could go to bed.

    [20.7]Call knew there was no point in arguing. [20.8]That was what Augustus wanted: argument. [20.9]He didn't really care what the question was, [20.10]and it made no great difference to him which side he was on. [20.11]He just plain loved to argue. [McMurtry, *Lonesome Dove*, p. 16]

Sentences (20.3)-(20.6) are Augustus's subjective sentences and (20.7)-(20.11) are Call's. Thus, (20.7) initiates a new POV. It is a private-state sentence and the SC, Call, is the experiencer of the private state. But passage (20) also shows that the SC of a private-state sentence need not be the experiencer. In (20.6), for example, "Call didn't like the conversation" is about a private state (Call not liking the conversation), but the SC of the sentence is Augustus, not Call. In the following subsections, we will consider factors that can indicate that it is *not* the experiencer who is the SC of a private-state sentence.

**8.3.1 Textual Continuity.** POV does not typically shift from one character to another without a paragraph break. Thus, the absence of a paragraph break suggests that a shift has not occurred. Consider the following schema, in which a subjective sentence $S$, whose SC is $X$, is followed, without a paragraph break, by a private-state sentence $P$ whose experiencer is a different character $Y$:

¶   sentence*    subjective-sentence-$S$   private-state-sentence-$P$
                  SC $= X$                 experiencer $= Y$

Character $Y$ is the experiencer of the private state that $P$ is about. If $Y$ were also the SC of $P$, then a shift would have occurred, from $X$'s POV to $Y$'s POV, without a paragraph break. The fact that no paragraph break appears—that is, that $P$ is in the continuing-subjective situation—suggests that $P$ continues $X$'s POV rather than initiating $Y$'s. When a private-state sentence appears in the continuing-subjective situation, therefore, the algorithm identifies the SC to be the last SC rather than the experiencer of the private state.

    The question of whether there is a psychological link between paragraph breaks and tracking POV has not been previously investigated. Stark (1987 and 1988) performed psychological experiments that showed that there is a significant correlation between paragraph breaks and discourse discontinuities, but the sorts of discontinuities she investigated did not include changes in POV. Nakhimovsky and Rapaport (1988) suggest that in narrative, paragraph breaks accompany changes in POV, but they did not investigate this hypothesis experimentally. In fact, we have performed psychological experiments (Bruder and Wiebe 1990 and forthcoming) that did establish such a link. Specifically, through manipulation of paragraph breaks in naturally occurring passages, the experiments showed that readers' interpretations of private-state sentences are influenced by paragraph breaks as we predicted on the basis of the algorithm.

**8.3.2 Subjective Elements.** Private-state sentences are pragmatically ambiguous as to whether they are private-state reports or represented thoughts.[4] Consider sentence (21):

(21) John knew Mary had the key.

Sentence (21) is about a private state:

$$\text{``John} \quad \text{knew} \quad \text{Mary had the key.''}$$
$$\downarrow \qquad \downarrow \qquad \downarrow$$
$$ps(p_1, experiencer_1, \quad attitude_1, \quad object_1)$$

Under a private-state *report* interpretation of (21), $p_1$ is not itself the object of some other private state. But under a represented thought interpretation of (21), $p_1$ *is* the object of some other private state $p_2$, the experiencer and attitude of which are implicit:

$$\text{``John knew Mary had the key.''}$$
$$\downarrow$$
$$ps(p_2, experiencer_2, \quad attitude_2, \quad object_2 = p_1)$$

To my knowledge, this ambiguity in the interpretation of private-state sentences and its importance in tracking POV have not been previously discussed in linguistics or literary theory. For example, Cohn (1978) says that represented thoughts can be distinguished from private-state reports by "the absence of mental verbs" in the former (p. 104).

The SC of a private-state report is always the experiencer of the private state. So, if some oracle were to inform you that (21) is a private-state report, you would then know that the SC is the experiencer of the private state (John). On the other hand, if the oracle were to inform you that (21) is a represented thought, you could not then identify the SC just by looking at the sentence alone. In fact, it is true of any represented thought without a narrative parenthetical, private-state sentence or otherwise, that you cannot identify the SC from the sentence itself. This is so regardless of whether or not the SC happens to be referred to in the sentence. Consider the following two sentences, which are represented thoughts from different pages of a short story ("The Garden Party" by Katharine Mansfield):

(22) Why couldn't she?
(23) What nice eyes he had, small, but such a dark blue!

As these sentences appear in the story, the SC of (22) happens to be the referent of "she" (corresponding to the conversational utterance, "Why can't I?"), but the SC of (23) is not mentioned in the sentence at all. Even though you know that (22) and (23) are represented thoughts, you need to consider the context to identify their SCs. Thus, if a private-state sentence $s$ contains some indication that $s$ is a represented thought, then the SC cannot be identified from $s$ itself, and, as discussed above in section 8.2, the expected subjective characters should be considered.

**Subjective elements** indicate that a sentence is a represented thought (this statement is qualified later in this section and in section 10 below). Subjective elements are linguistic elements that express emotions, uncertainty, evaluations, and other kinds of subjectivity (Banfield 1982) (they are discussed in detail below in section 9). Examples are evaluative terms such as 'the old bag' (Banfield 1982) and evidentials such as 'evidently' and 'apparently' (Doležel 1973).

In the following passage, a subjective element indicates that the SC of a private-state sentence is not the experiencer of the private state. At the start of the passage, Sandy and Dennys are (collectively) the last subjective character:

(24)
$^{24.1}$Japheth, evidently realizing that they were no longer behind him, turned around $^{24.2}$and jogged back toward them, seemingly cool and unwinded. [L'Engle, *Many Waters*, p. 24]

The subjective element 'evidently' in (24.1) indicates that the sentence is *not* a private-state report. That is, (24.1) is not a report that Japheth realizes that they are no longer behind him. Rather, Sandy and Dennys (the collective SC) ascribe this private state to him.

However, **subordinated** subjective elements, those *within* the scope of the private-state term, can appear in private-state reports. (This is one reason why I define private-state reports to be subjective.) Thus, they cannot be used to distinguish private-state reports from represented thoughts, and so cannot be used as evidence that the SC of a private-state sentence is not the experiencer. For example:

(25)
$^{25.1}$Ugh! she [the girl] thought. $^{25.2}$How could the poor thing have married him in the first place?
$^{25.3}$Johnnie Martin could not believe that he was seeing that old bag's black eyes sparkling with disgust and unsheathed contempt at him. [Caldwell, *No One Hears But Him*, pp. 98-99]

Sentence (25.3) is a private-state report and the experiencer *is* the SC (Johnnie Martin); this is so even though (25.3) contains the subjective element 'old bag' and even though there is an expected subjective character (the girl) when it is encountered. Because 'old bag' appears within the scope of the private-state term 'believe', it is not considered in identifying the SC. On the other hand, the subjective element 'evidently' in (24.1) is not in the scope of 'realizing' (i.e., it is **non-subordinated**); thus, it *can* be used as evidence that the SC is not the experiencer of the private state.

If a private-state sentence does not have a non-subordinated subjective element and does not appear in the continuing-subjective situation, then the algorithm identifies the SC to be the experiencer.

**8.3.3 Broadening and Narrowing of POV.** Recall that an experiencer, actor, SC, or expected subjective character may be more than one physical character (recall that these entities are represented as sets). A **broadening** of point of view occurs when a new subjective character is a superset of the old subjective character, and a **narrowing** occurs when a new one is a subset of the old one. One situation in which such changes occur is when the experiencer of a private-state report is such a subset or superset. One addition to the algorithm as described so far is needed to allow for this situation: for a private-state sentence in the continuing-subjective situation (without subjective elements that can be considered), if the experiencer is a superset or subset of the last subjective character, then it is the experiencer who the algorithm chooses to be the SC, rather than the last subjective character. For example:

(26)
$^{26.1}$In the clear late afternoon light they [Call and Augustus] could see all the way back to Lonesome Dove and the river and Mexico. $^{26.2}$Augustus regretted not tying a jug to his saddle—$^{26.3}$he would have liked to sit on the little hill and drink for an hour. [McMurtry, *Lonesome Dove*, p. 241]

As this passage appears in the novel, (26.1) is the subjective sentence of both Call and Augustus, but (26.2) is the subjective sentence only of Augustus; thus, point of view narrows upon (26.2). The above rule precludes the reading of a sentence such as (26.2) as the represented thought of the last subjective character (i.e., Call and Augustus in passage (26)). However, no examples of such a reading were found in the texts examined; note that the rule applies only if there are no subjective elements that can be considered in the sentence.

**8.3.4 Unspecified Experiencers.** Consider the following passage:

(27)
$^{27.1}$For the first time in his life, Sandy had a flash of gratitude that Dennys was not with him.
  $^{27.2}$Then anxiety surfaced. $^{27.3}$"Dennys—" [L'Engle, *Many Waters*, p. 9]

Sentence (27.2) is a private-state sentence; specifically, CHOSEN-STATE-OF-AFFAIRS(FEATURES (27.2)) = $ps$, where $ps$ is the private state that the private-state noun "anxiety" is about (see section 6.3.3 above). Both the experiencer of $ps$ and the SC of (27.2) happen to be the same character (Sandy). However, the SC of (27.2) cannot be identified from the sentence itself, because the experiencer of $ps$ is not mentioned in the sentence. (I call such experiencers **unspecified experiencers**; note that EXPERIENCER-OR-ACTOR-OF($ps$,FEATURES(27.2)) = {}.) Thus, for a private-state sentence with an unspecified experiencer, IDENTIFY-SC-FROM-THE-SENTENCE returns the empty set, and the algorithm goes on to consider expected subjective characters. In (27), for example, the algorithm correctly identifies the SC of (27.2) to be Sandy, the last subjective character.

**8.3.5 Summary.** Following is the algorithm for function IDENTIFY-SC-FROM-THE-SENTENCE (introduced in section 8.1 above). One of the arguments is a feature set, *featureSet*; we will use $s$ to refer to the sentence that *featureSet* is a feature set of (i.e., FEATURES($s$) = *featureSet*). With the exception of calls to previously mentioned functions, statements are given in English. Further, two cases discussed later (in sections 10 and 11) are not included (the complete version is given in appendix I).

IDENTIFY-SC-FROM-THE-SENTENCE (*featureSet, context*)
$soa \leftarrow$ CHOSEN-STATE-OF-AFFAIRS (*featureSet*)
**if** $s$ contains a narrative parenthetical **then**
    **return** the subject of the narrative parenthetical
**else if**
    1 (*soa* is a private state ) **and**
    2 % not an unspecified experiencer
      (EXPERIENCER-OR-ACTOR-OF(*soa, featureSet*) ≠ {}) **and**
    3 (There are no non-subordinated subjective elements in the sentence) **and**
    4:
        (a) ((TEXT-SITUATION-OF(*context*) ≠ continuing-subjective) **or**
        (b):
            (i) ((TEXT-SITUATION-OF(*context*) = continuing-subjective) **and**
            (ii): % broadening or narrowing of POV
                (1) ((EXPERIENCER-OR-ACTOR-OF(*soa*) ⊂
                    LAST-SC-OF (*context*) **or**
                (2) (EXPERIENCER-OR-ACTOR-OF (*soa*) ⊃
                    LAST-SC-OF (*context*))))
    **then return** EXPERIENCER-OR-ACTOR-OF(*soa*)

```
    else return {}
    end if
end if
```

### 9. Recognizing Subjective Sentences

#### 9.1 Introduction
We now turn to deciding whether or not a sentence is subjective in the first place. Authors could unambiguously mark each subjective sentence as subjective, by including a narrative parenthetical in each, for example. But suppose that a sentence $S$ that the author intends to be subjective appears in the continuing-subjective situation:

$$(i) \quad \P \quad \text{sentence}^* \quad \text{subjective-sentence} \quad \text{sentence-}S \quad \text{sentence}^* \quad \P$$
$$\text{SC} = X$$

A character's POV very often continues at least until the end of the paragraph. So, in schema $(i)$, $S$ and any sentences after $S$ until the paragraph break will very often be subjective sentences of $X$. Thus, the reader has a strong expectation that $X$'s POV will continue, so a weak hint that $S$ is subjective is sufficient for the reader to recognize that it is.

Now consider a text situation in which there has been a subjective sentence in the scene, but objective sentences and paragraph breaks have appeared since then:

$$(ii) \quad \text{start-of-scene} \quad \ldots \quad \text{subjective-sentence} \quad (\text{objective-sentence}^+ \; \P)^+ \quad \text{sentence-}S$$
$$\text{SC} = X$$

In $S$'s context in schema $(ii)$, $X$ is an expected subjective character. The reader expects $X$'s POV to be resumed, but not as strongly as the reader expects $X$'s POV to be continued in schema $(i)$, since the local context of $S$ in $(ii)$ is not subjective as it is in $(i)$. An unambiguous indication that $S$ is subjective is not necessary, but a stronger hint should be included than is sufficient in $(i)$.

The main sorts of "hints" of subjectivity that the algorithm considers are linguistic elements that potentially express subjectivity (**potential subjective elements**). Some of these are weaker hints than others, and many are usually subjective only in certain text situations. The algorithm, which, recall, tracks point of view on the basis of regularities, uses the text situation to decide whether an instance of one is indeed subjective.

"Subjective element" is the term I use for an instance of a potential subjective element that actually is subjective in the context of use. This term is borrowed from Banfield (1982), but redefined; Banfield uses it to refer only to linguistic elements that are *always* subjective.

Section 9.2 identifies a number of potential subjective elements, and section 9.3 specifies how the algorithm uses them and other information to recognize subjective sentences.

#### 9.2 Potential Subjective Elements
Previous work in linguistics and literary theory noted the presence in subjective sentences of many (but not all) of the potential subjective elements considered by the algorithm. However, with the exception of two of the elements, the perfective and progressive aspects (see below), previous work did not address the problem that many of the elements are only potentially subjective, and can also appear in objective sentences. Further, Wiebe 1990 contains an extensive, detailed catalogue of the potential subjective elements (specified mainly in terms of syntactic and semantic categories presented in Quirk et al. 1985); such a catalogue did not previously exist.

Most of the potential subjective elements are lexical. But it is not words and phrases themselves that are potential subjective elements, but rather words and phrases used with particular meanings. For example, 'poor' is a potential subjective element only with its evaluative meaning, as in "Poor John was sick", but not with its non-evaluative meaning, as in "John was poor" (Banfield 1982; the evaluative meaning of 'poor' is one of the elements that Banfield argues is always subjective).

Tables 2 and 3 list some potential subjective elements, giving very brief characterizations. For further details, see Wiebe 1990. All of the citations in tables 2 and 3 are with respect to the linguistic categories of the elements. Some of those who discuss the appearance of the elements in subjective sentences are as follows: Banfield (1982) discusses (1), (2.1)-(2.3), (4), (5), and (12); Doležel (1973) discusses (1), (2.1)-(2.3), (5), (6.1), (6.2), and (8); Brinton (1980) discusses (9) (she shows that a simile can be a marker of represented perception), and (12); Ehrlich (1990) discusses (11), and Ehrlich (1987) discusses (12).

The past perfective is potentially subjective simply because a character can reflect on what occurred (or might have occurred) in the past. However, as discussed by Ehrlich (1990) and many others, the narrative past may be expressed by the simple-past tense in the midst of a subjective context; detecting simple-past references to the past is not addressed in this work.

---

1 Exclamations, such as (25.1), and direct questions, such as (25.2)

---

2 Elements that express evaluation or judgement

    2.1 Adjectives such as 'awful' in (4.6) and 'poor' in (25.2)

    2.2 Nouns such as 'old bag' in (25.3)

    2.3 Adverbs such as 'oddly' and 'incredibly'

    2.4 Auxiliary verbs and phrases that express judgments of obligation, such as 'had better', 'ought to', 'should', and 'be supposed to'

    2.5 Adverbs such as 'scarcely' and 'hardly' (when used as minimizer subjuncts (Quirk et al. 1985)), as in "She could hardly be expected to live there"

---

3 Elements that express a lack of knowledge

    3.1 Subordinators such as 'whoever' and 'whatever', when used in reference to particular individuals, as in "Whatever it was, it had flown by quickly"

    3.2 Adjectival phrases such as 'some kind of', when used in reference to particular individuals, as in "The object in her hand was some kind of weapon"

---

Table 2
Some potential subjective elements.

### 9.3 Recognizing Subjective Sentences

Subjective elements are important for recognizing represented thoughts and perceptions, not private-state reports (recall, in fact, that a non-subordinated subjective element is evidence that a private-state sentence is *not* a report). That is, they are important for recognizing subjective sentences whose subjective characters are to be identified from the

4 Sentence fragments, such as (30.6)

5 Kinship terms, such as 'Dad' and 'Aunt Margaret'

6 Evidentials, which, in the broadest sense, qualify the information conveyed by a statement (Chafe 1986)
  6.1 Evidentials that express certainty or uncertainty, such as 'surely' and 'might'
  6.2 Evidentials that express certainty or uncertainty and also that one's knowledge is based partly on evidence. Examples are 'evidently', 'seemingly', 'must have', 'appear to be', 'as if', 'as though', and 'look', as in "He looked like he might cry"
  6.3 Hedges, e.g., adverbs such as 'more or less' and 'sort of' when used as modifiers of adjectives and adverbs, as in "It was more or less green", or as adverbials (Quirk et al. 1985), as in "The man more or less held a large stretch of the border"
  6.4 Evidentials that address expectations
    6.4.1 Signal that expectations *have* been met, such as 'of course' (when used as an emphasizer subjunct (Quirk et al. 1985)) as in "John of course sat down"
    6.4.2 Signal that expectations have *not* been met. Examples are adverbs such as 'just', 'merely', and 'only' (when used as attitude diminishers (Quirk et al. 1985)), as in "He just sat and drank" (it was expected that he would do something "more" than sit and drink)

7 Adverbials that are *conjuncts*, which connect units of discourse (Quirk et al. 1985) (i.e., *cue phrases*; Reichman 1985, Grosz and Sidner 1986, Cohen 1987). Examples are 'first', 'in addition', 'for instance', 'on the other hand', 'after all', 'anyway', and 'yet' as in "Yet, they were the pride of the family"

8 Conditional clauses

9 Comparative 'like', as in "They followed her like acolytes behind a goddess"

10 Habitual sentences, such as "Gus himself often joked about it"

11 The past perfective, but only in the main verb phrase

12 The progressive, but only in the main verb phrase

**Table 3**
Some potential subjective elements (continued).

context, rather than from the sentence itself.

My examination of novels and short stories suggests the following (we are currently performing psychological experiments investigating the aspects of the algorithm discussed in this section): (1) Two potential subjective elements, the past perfective and the progressive, can typically serve only to continue a character's POV and only within a paragraph (see Ehrlich 1987 for an analysis of why this is so for the progressive); (2) stronger ones can continue a character's POV after a paragraph break, or resume a character's POV within a paragraph; (3) still stronger ones, such as evidentials and sentence fragments, can resume the last subjective character's POV or initiate the last active character's just as long as they are expected subjective characters; and (4) the strongest subjective elements, such as exclamations and questions, are always subjective, even when there is not an expected subjective character to whom to attribute the sentence. The sets of text situations corresponding to (1)-(4) are:

$1_{ts}$ {continuing-subjective}

$2_{ts}$ {broken-subjective, interrupted-subjective}

$3_{ts}$ {presubjective-active, postsubjective-nonactive, postsubjective-active}

$4_{ts}$ {presubjective-nonactive}

Expectations for a subjective sentence are strongest in situation ($1_{ts}$) and weakest in situation ($4_{ts}$), so the algorithm takes even the weakest potential subjective elements to be subjective in ($1_{ts}$), but only the strongest ones to be subjective in ($4_{ts}$). In general, each potential subjective element *pse* is associated with a set of text situations $t$ such that the algorithm interprets *pse* to be subjective iff the current text situation is in $t$. There is an $i_{ts}$ ($1 \leq i \leq 4$) such that $t$ contains the situations in $1_{ts}$ through $i_{ts}$ but not those in $i_{ts} + 1$ through $4_{ts}$. We shall say that *pse* is associated **at the highest level** with the situations in $i_{ts}$.

In addition to potential subjective elements, there is another source of information the algorithm considers: the type of state of affairs the sentence is about. First, private-state **action** sentences can be subjective; see section 11 below. Second, private-state sentences are usually subjective (we are not considering objective private-state sentences in this paper; see Wiebe 1990 and footnote 4). Third, a nonprivate-state sentence in the continuing-subjective situation usually continues the subjective context. For example:

(28)
$^{28.1}$Lorena didn't like it that Gus acted like Jake wasn't much. $^{28.2}$He had a reputation for being a cool man in a fight. [McMurtry, *Lonesome Dove*, p. 190]

Sentence (28.1) is Lorena's subjective sentence, and (28.2), a nonprivate-state sentence, continues her subjective context. In the continuing-subjective situation, therefore, the algorithm interprets a nonprivate-state sentence to be subjective.[5]

### 9.4 Examples and Discussion
Consider the following passage:

(29)
$^{29.1}$Call had heard from someone that she had been raised rich, in the East, with servants to comb her hair and help her into her shoes when she got up. $^{29.2}$It might just have been a story— $^{29.3}$it was hard for him to imagine a grownup who would need to be helped into their own shoes— $^{29.4}$but if even part of it was true she had come a long way down.

> [29.5]Ned Spettle had never got around to putting a floor in the shack of a house he built. [29.6]His wife was rearing eight children on the bare dirt. [29.7]He had heard it said that Ned had never got over the war, which might have explained it. [McMurtry, *Lonesome Dove*, p. 176]

All of these sentences are Call's subjective sentences. Thus, the text situation is continuing-subjective after each of them. Sentences (29.1), (29.3), and (29.7) are Call's private-state reports, and (29.2) and (29.4) contain potential subjective elements that are associated with other situations than merely the continuing-subjective one ('might' in (29.2) and 'if even' in (29.4)). The interesting sentences are (29.5) and (29.6), since they contain potential subjective elements that are associated only with the continuing-subjective situation (the past perfective in (29.5) and the progressive in (29.6)). (These sentences express Call's reasons for his belief, expressed in (29.4), that "she had come a long way down".)

In the following passage, a subjective element appears in a situation other than continuing-subjective. The situation is continuing-subjective at the beginning, and Sandy is the last subjective character.

> (30)
> [30.1]The eyes were an incredibly bright blue, like the sea with sunlight touching the waves.
> [30.2]Lemech greeted him respectfully. [30.3]"Adnarel, we thank you."
> [30.4]Then he said to Sandy, "The seraph will be able to help you. Seraphim know much about healing."
> [30.5]So this was a seraph. [30.6]Tall, even taller than the twins. [L'Engle, *Many Waters*, p. 39]

Sentence (30.1) continues Sandy's subjective context, because it contains the subjective elements 'incredibly' and comparative 'like'. Sentences (30.2)-(30.4) are objective, and a paragraph break appears before (30.5), so the situation at the beginning of (30.5) is postsubjective-nonactive (one of the situations in ($3_{ts}$)) and the last subjective character, Sandy, is an expected subjective character. The algorithm is able to recognize that (30.5) is Sandy's subjective sentence, because it contains 'so' used as a conjunct, which is subjective as long as there is an expected subjective character.

Consider passage (19) (reprinted here); recall that it is the beginning of a novel, so there isn't an expected subjective character when it is encountered.

> (19)
> [19.1]Captain Scalawag's treasure! [19.2]It was the first thing Pete thought of when he woke up. [Lorimer, *The Mystery of the Missing Treasure*, p. 1]

Since an exclamation is subjective in any situation, the algorithm is able to recognize that (19.1) is subjective.

In the following passage, in contrast, potential subjective elements that are *not* necessarily subjective appear when there isn't an expected subjective character. This passage is of the type that Banfield has characterized as having an **empty center** (Banfield 1987);[6] it describes perceptions and impressions that one would have if observing the scene, but no character is present to whom to attribute them. There is an expected subjective character at the beginning of the passage, but a scene break appears after the third sentence. (The blank lines after (31.3) and (31.12) appear in the original. The sentences following '—' are a kind of unuttered quoted speech.)

(31)

> [31.1]"We're coming," Oholibamah said. [31.2]And they hurried toward the central section of the oasis, where Noah's vineyards were, and his grazing grounds, and his tents. [31.3]And where Dennys was waiting for them.
>
> [31.4]The moon set, its path whiter than the desert sands dwindling into shadow. [31.5]The stars moved in their joyous dance across the sky. [31.6]The horizon was dark with that deep darkness which comes just before the dawn.
> [31.7]A vulture flew down, seemingly out of nowhere, stretching its naked neck, settling its dark features.
> [31.8]—Vultures are underestimated. Without us, disease would wipe out all life. We clean up garbage, feces, dead bodies of man and beast. We are not appreciated.
> [31.9]No sound was heard [31.10]and yet the words seemed scratched upon the air.
> ...
>
> [31.11]The twelve oddly assorted creatures began to position themselves into a circle.
> [31.12]The nephilim.
>
> [31.13]Oholibamah lay in Japheth's arms on a large, flat stone a short walk into the desert. [L'Engle, *Many Waters*, pp. 118-119]

There are scene breaks after (31.3) and after (31.12). Between the breaks, there are no private-state sentences or narrative parentheticals, and, since none of the creatures in the scene has been the SC, there are no sentences with active characters (this is true for the elided sentences as well). Thus, if none of the potential subjective elements are subjective, then there are no expected subjective characters from (31.4) to (31.12). There are strong potential subjective elements that would be subjective if there were an expected subjective character—'seemingly' in (31.7), 'sound' in (31.9), 'yet' used as a conjunct and 'seemed' in (31.10), 'oddly' in (31.11), and a sentence fragment in (31.12). However, since these are associated at the highest level with the situations in ($3_{ts}$), and not with the presubjective-nonactive situation, the algorithm correctly does not interpret them to be subjective.

There are three things to note about passages such as (31). First, the potential for an overt narrator to appear is strong when there isn't an expected subjective character and a strong element such as 'seemingly' appears; my restriction to texts that do not have overt narrators allows the algorithm to exclude this possibility from consideration. Second, the algorithm does not revise its decision as to whether a sentence is subjective in light of later sentences. However, one could imagine a sentence within passage (31) that might cause the reader to decide that earlier sentences were actually subjective. For example, the following sentence, inserted after (31.10), would suggest this: "Dennys was mystified by the spectacle." The algorithm would interpret this sentence to be subjective; this would affect its interpretation of the *remainder* of the passage—the algorithm would interpret the potential subjective elements in (31.11) and (31.12) to be subjective— but it would not affect its interpretation of earlier sentences. Third, to be conservative, only Banfield's emotive and evaluative subjective elements, which must be understood to express someone's emotions or evaluations, are associated with the presubjective-nonactive situation ($4_{ts}$). As mentioned above in section 8.2.4 the number of sentences that appear in this situation is relatively small, and, of those that do, many are private-state reports or have narrative parentheticals. It may be that some of the potential subjective elements associated at the highest level with the situations in ($3_{ts}$) should also be associated with the one in ($4_{ts}$); the appearance of the relevant kind of subjective sentence in this situation was too rare in the texts examined to decide this.

Because there are thirty-four classes of potential subjective elements, the majority of which have multiple members (Wiebe 1990), a significant number of each potential subjective element in each situation was not found. The association of elements with text situations is based on the examples that were found, and, for the ones that did not appear very often in the texts examined, on my judgments as to which of the ones that did appear often are closest to them in strength. Psychological experiments (with Gail Bruder) of this aspect of the algorithm are underway. We plan to revise the association of potential subjective elements with text situations as needed in light of the results.

### 9.5 Quoted Speech

In quoted speech, there are two points of view: the point of view taken by the quoted string—the speaker's—and the one taken by the discourse parenthetical, which may be objective or a character's.[7] It is the point of view taken by the discourse parenthetical that concerns us here: the speaker's point of view is not directly presented by a quoted string, as the subjective character's is by a subjective sentence, but is conveyed indirectly through a communicative act. Quoted speech is a major way of communicating the beliefs, intentions, etc., of characters who are not to become the SC; merely the fact that what a character says expresses her point of view should not lead the reader to anticipate later subjective sentences of that character, as the reader does after a subjective sentence.

Therefore, the algorithm considers subjective elements that appear in discourse parentheticals, but not those within quoted strings. For example:

> (32)
> "I'll talk to Amy," Daddy said, "and make sure she behaves herself."
> [Sachs, *Amy and Laura*, p. 100]

The subjective element 'Daddy' in the discourse parenthetical is attributed to an expected subjective character, Laura. In contrast, the algorithm interprets the following sentence from passage (20) to be objective, even though there is a question in the quoted string:

> "Drown me?" Augustus said.

### 10. A Return to Private-State Sentences

In contrast to what was implied above in section 8.3.2, there are some subjective elements that do not suggest that a private-state sentence is a represented thought, even when they are non-subordinated. First, Cohn (1978) shows that private-state reports[8] do not always report private states experienced specifically at the current moment in the story, but instead have "almost unlimited temporal flexibility" (p. 34). One consequence of this is that private-state reports can be habitual. Second, Cohn (1978) also shows that private-state reports can employ simile.[9] Thus, comparative 'like' can appear non-subordinated in private-state reports. For example:

> (33)
> His [Sandy's] head began to swell, to be filled with hot air like a balloon,

so that he was afraid he was going to float off into the sky. [L'Engle,
*Many Waters*,
p. 27]

Finally, some intensifier adverbs, when non-subordinated in a private-state report, simply indicate the degree to which the private state is experienced. An example is 'hardly', as in:

(34)
Sandy, his flannel shift still draped over his head, was hardly aware that he was supporting his brother. [L'Engle, *Many Waters*, p. 27]

Sentence (34), as it appears in the novel, is Sandy's private-state report.

Given these observations, we need to revise the algorithm as presented so far: the algorithm does not consider the above types of subjective elements when it decides who the SC of a private-state sentence is, even when they are non-subordinated to the private-state term.

**11. Private-State-Action Sentences**

A **private-state action** is an action from which a private state can be inferred. Examples are looking, glancing, sighing, frowning, smiling, and shivering. In contrast to a private-state report such as "She was unhappy", the sentence "She frowned" narrates a private-state action from which unhappiness or displeasure can be inferred, but does not directly report the character's private state.[10] In a given context, the private state that can be inferred may or may not be significant for tracking point of view. It is significant in the following passage:

(35)
$^{35.1}$Zoe looked at the notebook. $^{35.2}$On the first page Joe had written WAR WORK in large block letters in red and blue crayon. $^{35.3}$On the next page he had written the date $^{35.4}$and under it all about seeing Miss Lavatier's boyfriend. [Oneal, *War Work*, p. 47]

As this passage appears in the novel, (35.2)-(35.4) continue the subjective context established by (35.1)—they are subjective sentences presenting what Zoe sees. Interpreting (35.1) to be subjective, the algorithm is able to recognize that (35.2)-(35.4) are also subjective, because the past perfective is a subjective element in the continuing-subjective situation, and it is able to determine that the SC is Zoe, because she is the last subjective character. On the other hand, a private-state action sentence might *not* be the subjective sentence of the actor; an example is (36.1):

(36)
$^{36.1}$Japheth looked at them. $^{36.2}$"You are flushed. And wet." $^{36.3}$He himself did not seem to feel the intense heat. [L'Engle, *Many Waters*, p. 20]

As this passage appears in the novel, (36.3) is the subjective sentence of the last subjective character, Sandy and Dennys. If (36.1) were Japheth's subjective sentence, then it would be to Japheth that the subjective element 'seem' in (36.3) would be attributed, rather than to Sandy and Dennys.

Like quoted speech, a private-state-action sentence is a way to communicate something about the consciousness of a character who is not to become the SC. The reader infers from (36.1) that Japheth sees the people he is looking at; however, there are no subsequent subjective sentences about what he sees, such as the sentences in (35) that show what Zoe sees.

The first decision to be made when one encounters a private-state-action sentence is whether it should be treated as a private-state sentence or as an action sentence. Whether the sentence is subjective and, if so, who the SC is then depend upon the factors already presented.

The algorithm's treatment of these sorts of sentences is based on the observation that a more direct appeal to a character's consciousness, such as a private-state report or a narrative parenthetical, is usually used to establish a character as an SC for the first time. Thus, the actor of a private-state-action sentence that *is* the actor's subjective sentence has usually been the SC before. While consistency in the interpretation of a passage with this sort of sentence must be supported by other factors, this regularity is a strong one in the texts examined.

Thus, if the actor has been a SC, then the algorithm treats a private-state-action sentence *s* as it would have treated a private-state sentence in the same context; otherwise, it treats *s* as it would have treated an action sentence in the same context (see Wiebe 1990 for illustrations of the various consequences of treating *s* one way or the other).

## 12. Tests of the Algorithm

This section summarizes the tests of the algorithm that have been performed.

First, the algorithm was hand-simulated on over 700 pages (roughly 17,500 sentential input items) from seven novels that represent a range in the number of different subjective characters they contain. Given the large amount of data and the preprocessing requirements of running the algorithm, the purpose of this test was not to compile statistical measures, but rather to find out what kinds of exceptions occur. Generally, point of view is manipulated in these novels as expected by the algorithm. Of the exceptions, many can be attributed to problems that are specifically not addressed in this work, such as how the spatial and temporal points of view affect the psychological one (discussed in the next section), how point of view is manipulated in relatively rare situations, such as the very beginning of a novel, and what constitutes a "significant" subjective context for the purpose of interpreting private-state-action sentences (discussed in appendix III). (The classes of errors and examples of them are given in Wiebe 1990.)

In order to obtain more specific numeric results, the algorithm was tested, and results tabulated, for a more modest amount of text (32 pages with 900 sentential input items). The results are positive, and are given in appendix III. Fully testing the algorithm would require a much larger corpus, in which a significant number of each of the possibilities arises. Such testing would be a major endeavor in itself.

Finally, as mentioned in section 8.3.1, we have performed psychological experiments that showed that readers' interpretations of private-state sentences are affected by paragraph breaks as we predicted on the basis of the algorithm (Bruder and Wiebe 1990 and forthcoming). We are continuing this line of research with psychological experiments of other aspects of the algorithm.

## 13. Relation to Anaphora Resolution

A question likely to arise in the reader's mind is how tracking POV and anaphora resolution are related. Anaphora resolution is necessary for tracking POV (the actors and experiencers of states of affairs must be known; see item (4) of section 6.1). But it is certainly not sufficient, and an additional mechanism is needed. A case that clearly illustrates this is a private-state sentence in which the experiencer is referred to with a pronoun, such as:

> (37) He hated holidays.

Resolving the pronoun is not sufficient for identifying the SC of (37), since the SC may or may not be the referent of 'he' (see section 8.3.2).

However, the pipeline architecture of the algorithm is not realistic. Almost certainly, POV affects anaphora resolution, and also recognizing the discourse structure of the text. Specific issues I am exploring are discussed in the next section.

## 14. Directions for Future Research and Conclusion

One direction for future research is investigating the interactions of different points of view. A large class of exceptions to the algorithm can be attributed to interactions between the spatial and temporal points of view and the psychological one. For example, there are spatial and temporal discontinuities other than scene breaks after which a character should no longer be expected: the current "here" and "now" in the story may shift away from the character, or the character may leave the scene while the "here" and "now" remain unchanged. These situations must be distinguished from the situation, occurring with represented thoughts, in which there is a "projected here" and a "projected now" of events that are being thought about, and from the situation in represented perception in which there is a "projected here" of events being perceived (Bruder et al. 1986, Rapaport et al. 1989ab).

An important area of future research is investigating interactions among tracking POV, recognizing discourse structure, and anaphora resolution. I am currently focusing on discourse structure within subjective contexts. We can view a subjective context as presenting a sequence of private states, where the experiencers and attitudes of some are only implicit (in the case of represented thoughts). In addition to discourse relations among sentences as wholes, there can be discourse relations among *objects* of private states, even a hierarchical structure involving several such objects. As a concrete example, the contrast signaled by the cue-phrase 'but' appearing in a represented thought may be between a represented thought and the *object* of a previous private-state report. One example of a potential effect on pronoun resolution: if a sentence $s$ is to be interpreted as a represented thought, and is to be incorporated into a discourse-segment structure among previous private-state objects, then pronouns in $s$ can be resolved against the focus space(s) corresponding to the private-state-object segments (as in Grosz and Sidner's (1986) theory). Of course, determining whether or not $s$ is to be so interpreted is a difficult problem. But such interactions among POV, discourse structure, and anaphora resolution might be usefully cast as constraints, with evidence regarding interpretation with respect to one limiting the options to be pursued for the others.

The psychological POV is related to other pragmatic and discourse phenomena. Subjective contexts are paragraph-level analogues of opaque contexts in belief reports (Wiebe (1991) specifically addresses this issue). In addition, the discourse phenomena identified by Fauconnier (1985) are similar to the psychological point of view. Just as a private-state report can begin a discourse segment in which subsequent sentences are understood to continue a character's point of view, an adverbial such as "in 1969" can begin a discourse segment in which subsequent sentences are understood to refer to events that occurred in 1969, even though the date is not subsequently mentioned. Or, consider discussing someone else's work, say Smith's, in a research paper or text book. After an initial reference to Smith's work, you may go on to describe his or her theory without explicitly saying in each sentence that you are doing so (with a locution such as "Smith shows that", "In Smith's theory", "In Smith's algorithm", or "According to Smith") (William J. Rapaport, personal communication). Along with subjective contexts, an NLU system must be able to recognize such discourse phenomena in order to recover information implicitly communicated in the discourse. A detailed investigation of one of them suggests directions for investigating the others. The one investigated here, subjective contexts in particular kinds of texts, is a good one to investigate because it is possible to constrain the problem, because there are so many prototypical instances in those kinds of texts, and because there is a great deal of previous work in linguistics and literary theory to build upon.

In conclusion, this paper presented an algorithm for tracking the psychological point of view in third-person fictional narrative text. The algorithm is based on regularities, found by extensive examination of naturally occurring narratives, in the ways that authors manipulate point of view. The algorithm has been implemented, preliminary empirical studies have been performed, which support the algorithm, and psychological experimentation is continuing. This is the first detailed computational approach to the problem of tracking the psychological point of view.

**Acknowledgments.** I wish to thank the members of the SUNY Buffalo Graduate Group in Cognitive Science and the SNePS Research Group for many discussions and ideas, and William J. Rapaport, Graeme Hirst, Peter Heeman, Russ Greiner, the anonymous reviewers, and especially Diane Horton for helpful comments on earlier drafts of this paper. This research was supported in part by National Science Foundation grants IST-8504713 and IRI-8610517, and the preparation of this paper was supported in part by the Natural Sciences and Engineering Research Council of Canada.

**Texts Cited**

**Notes**

∗ Department of Computer Science, New Mexico State University, Box 30001/Dept. CS, Las Cruces, NM 88003-0001; wiebe@nmsu.edu

1 My wording in this paper often attributes agency to sentences. I might say, for example, that a sentence states something, communicates something, initiates a new POV, or refers to someone. Sidner (1983) and Webber (1983) object specifically to using a noun phrase as the agent of the verb 'refer', since it is the writer who is doing the referring, not the noun phrase. I do not disagree—my wording is for convenience only.

2 I am indebted to Stuart C. Shapiro for suggesting the names.

3 The fact that a sentence $r$ is habitual is part of FEATURES($r$). The system demonstrated in appendix II relies on adverbials to decide whether or not a sentence is habitual. However, an adverbial is not necessary for a simple-past narrative sentence to be habitual.

4 A private-state sentence may also be objective. An example is a simple-past sentence with a negated factive term and a propositional object, such as "John did not know that Mary was in the next room". This sentence cannot be John's subjective sentence; it is either objective or the subjective sentence of someone else. Due to space limitations, how the algorithm recognizes and processes such sentences is not discussed in this paper; see Wiebe and Rapaport 1988 and Wiebe 1990.

5 Notice that there are three things that are taken as evidence that a sentence is subjective only in the continuing-subjective situation: being a nonprivate-state sentence, being in the perfective, and being in the progressive. Caenepeel (1989), in work done simultaneously, analyzed aspect and perspective linguistically, and reached similar conclusions with respect to these three types of evidence. Caenepeel suggests that states appearing in what I call the continuing-subjective situation continue the current POV. Her notion of state includes sentences in the perfective and progressive aspects, regardless of the type of state of affairs that the sentence is about. It might be desirable to revise this aspect of the algorithm on the basis of her work, which focused on aspect, to arrive at a more general treatment of these three kinds of evidence in this situation.

6 Banfield 1987 extends the definition of subjective sentences given in Banfield 1982 to include these kinds of sentences; our interest is to recognize *character's* subjective sentences (for the reasons given in section 3), so we adopt the earlier definition.

7 See Banfield 1982 for a principled account of the relationship between the points of view of a quoted string and its discourse parenthetical.

8 Cohn's term for a private-state report is **psycho-narration**.

9 Cohn's term for such reports is **psycho-analogy**.

10 Brinton (1980) notes that perceptions may be reported from an "outer perspective" with terms such as 'gaze' and 'look' or from an "inner perspective" with terms such as 'see' and 'hear' (p. 370-371), but she does not address the significance of this difference for tracking point of view.

## Appendix I: The Algorithm

### AI.1 Introduction

The following aspects of the algorithm as it is presented in Wiebe 1990 are not addressed in this paper or included in the version of the algorithm given below.

1. In addition to the state-of-affairs types listed in section 6.3.1, there is an additional one, a **seeming state**. For example, "The man seemed tired to John" is about a seeming state. Such states are treated as private states.

2. The types of constituents, mentioned in section 6.3.3, in which private-state terms are not considered by the algorithm when choosing a state of affairs to consider. The example given in section section 6.3.3 was a manner adverbial.

3. Objective private-state sentences (see footnote 4).

4. When the SC cannot be identified when the sentence appears, identifying it after later sentences are processed (in section 8.2.4).

5. A listing of all of the potential subjective element categories (section 9.2) and the text situations with which each is associated (section 9.3).

6. A listing of all of the subjective elements that are not considered when they appear in private-state sentences (section 10).

This appendix is organized as follows. The preprocessing functions and operations on their ranges are specified in section AI.2; intermediate-level functions are given in AI.3; and the high-level functions are given in AI.4.

Function names are given in capital letters; parameter and variable names are given in italics; and comments are preceded by percent signs. The type of result returned by a function is given in the function heading following a colon. Preconditions are preceded by **Given**.

Only specifications are given for the functions in AI.2. These consist of preconditions, preceded by **Given**, and a specification of the result, preceded by **Returns**.

### AI.2 Preprocessing Functions.

• **Function Item and type-predicates on input items.**
ITEM($t$, $i$): input item

**Returns:** The input item at position *i* in text *t*.

SENTENCE(*item*): boolean
**Returns:** true iff input item *item* is a sentence.

PARAGRAPH-BREAK(*item*): boolean
**Returns:** true iff input item *item* is a paragraph break.

SCENE-BREAK(*item*): boolean
**Returns:** true iff input item *item* is a scene break.

- **Function Features and operations on feature sets.**
FEATURES(s): feature set
**Returns:** The feature set for sentence *s*.

Following are the operations on feature sets, however they may be implemented.

- **Item (1) of section 6.1.**

POTENTIAL-SUBJECTIVE-ELEMENTS(*featureSet*): set of potential subjective elements
**Returns:** the set of potential subjective elements in *featureSet*.

TYPE-OF-PSE(*pse, featureSet*): potential subjective element type (e.g., habitual, comparative-'like')
**Given:** $pse \in$ POTENTIAL-SUBJECTIVE-ELEMENTS(*featureSet*)
**Returns:** The type of potential subjective element that *pse* is

- **Item (2) of section 6.1.**

CLAUSES (*featureSet*): set of clauses
**Returns:** the set of clauses in *featureSet*.

SUBORDINATING-CLAUSES(*clause, featureSet*): set of clauses
**Given:** clause $\in$ Clauses (featureSet)
**Returns:** The set of clauses to which *clause* is syntactically subordinated.

STATES-OF-AFFAIRS (*featureSet*): set of states of affairs
**Returns:** the set of states of affairs in *featureSet*.

TYPE-OF-SOA (*soa, featureSet*): one of private-state-action, action, private-state, nonprivate-state
**Given:** $soa \in$ STATES-OF-AFFAIRS(*featureSet*)
**Returns:** The type of state of affairs that *soa* is.

SOA-OF-CLAUSE (*clause, featureSet*): state of affairs
**Given:** $clause \in$ CLAUSES(*featureSet*)
**Returns:** The state of affairs that *clause* is about

- **Item (3) of section 6.1.** These are the only functions that access information about the head noun of the subject of the main clause. A feature set must contain an indication as to whether or not this noun is about a private state, and, if this indication is true, the feature set must contain the state of affairs that this noun is about. But if this indication is false, even if the noun is about another kind of state of affairs, the feature set need not contain that state of affairs.

SOA$_{hn}$-Is-About-A-Private-State(*featureSet*): boolean
**Returns:** True iff the head noun of the subject of the main clause is about a private state.

SOA$_{hn}$(*featureSet*): state of affairs
**Returns:** if SOA$_{hn}$-Is-About-A-Private-State(*featureSet*), then returns the
  member of States-Of-Affairs (featureSet) that the head noun of the subject
  of the main clause is about. Otherwise, returns nil.

- **Item (4) in section 6.1.**

Experiencer-Or-Actor-Of (*soa, featureSet*): set of characters
**Returns:** {}, if $soa \notin$ States-Of-Affairs(*featureSet*).
  Otherwise, returns the experiencer or actor of *soa* (possibly the empty set).

- **Item (5) in section 6.1.**

Narrative-Parenthetical (*featureSet*): boolean
**Returns:** true iff the sentence contains a narrative parenthetical.

Subject-Of-Narrative-Parenthetical(*featureSet*): set of characters
**Given:** Narrative-Parenthetical (*featureSet*)
**Returns:** The subject of the narrative parenthetical in the sentence.

- **Item (6) in section 6.1.**

Is-In-The-Simple-Past(*clause, featureSet*): boolean
This is a pattern for other functions used by Active-Character-Of in AI.4.
**Given:** *clause* $\in$ Clauses (*featureSet*)
**Returns:** True iff the main verb phrase of *clause* is in the simple past.

Is-PSE-Subordinated-To-SOA (*pse, soa, featureSet*): boolean
**Given:** *pse* $\in$ Potential-Subjective-Elements (*featureSet*) and *soa* $\in$ States-Of-Affairs(*featureSet*)
**Returns:** False, if *soa* = SOA$_{hn}$(*featureSet*). Otherwise, let *clause* be the clause in Clauses(*featureSet*)
  such that SOA-Of-Clause(*clause, featureSet*) = *soa*, and let *l* be the lexical item(s)
  according to which it was determined that *clause* is about *soa*.
  Then, returns true iff *pse* is syntactically subordinated to *l*.

### AI.3 Intermediate-Level Functions.

These functions are used in many of the higher-level ones given in AI.4.

- **Context-Access Functions.**

Recall the context-access functions Last-SC-Of, Last-Active-Character-Of, Previous-SCs-Of, and Text-Situation-Of, introduced in section 6.2. Only the first is given here; the others are the obvious similar ones.

Last-SC-Of
    (*context* = $\langle$*lastSC, lastActiveCharacter, previousSCs, textSituation*$\rangle$):
    set of characters
**return** *lastSC*

- **Function To-Be-Treated-As-A-Private-State.**

To-Be-Treated-As-A-Private-State(*soa, featureSet, context*): boolean

**Given:** $soa \in$ STATES-OF-AFFAIRS($featureSet$) **and** TYPE-OF-SOA($soa, featureSet$) = private-state-action
**return** true iff EXPERIENCER-OR-ACTOR-OF($soa, featureSet$) $\subseteq$
   PREVIOUS-SCS-OF($context$)

• **Function Chosen-State-Of-Affairs and auxiliary functions.**
CHOSEN-STATE-OF-AFFAIRS($featureSet, context$): state of affairs
% Let $s$ be the sentence such that FEATURES($s$) = $featureSet$. Then,
% CHOSEN-STATE-OF-AFFAIRS($featureSet$) is the state of affairs that
% the algorithm will consider for $s$.
% The specification in section 6.3.3 does not mention private-state actions
% because they are not discussed until the end of the paper. However, they are included here.
$soa_{main} \leftarrow$ SOA-OF-CLAUSE(MAIN-CLAUSE($featureSet$), $featureSet$)
**if**
 1 (TYPE-OF-SOA($soa_{main}, featureSet$) = private-state) **or**
 2:
  (a) ((TYPE-OF-SOA($soa_{main}, featureSet$) = private-state-action) **and**
  (b) (TO-BE-TREATED-AS-A-PRIVATE-STATE($soa_{main}, featureSet, context$))) **then**
 **return** $soa_{main}$
**else if** SOA$_{hn}$-Is-About-A-Private-State($featureSet$) **then**
 **return** SOA$_{hn}$ ($featureSet$)
**else if** CANDIDATE-SUBORDINATED-CLAUSES($featureSet, context$) $\neq$ {} **then**
 **return**
  SOA-OF-CLAUSE(ONE-OF (CANDIDATE-SUBORDINATED-CLAUSES($featureSet, context$)),
    $featureSet$)
**else return** $soa_{main}$
**end if**

MAIN-CLAUSE ($featureSet$): clause
**return** the clause $c$ such that $c \in$ CLAUSES($featureSet$) and $Subordinating$-$Clauses$($c, featureSet$) = {}.

CANDIDATE-SUBORDINATED-CLAUSES ($featureSet, context$): set of clauses
**return** the set of all $c$ such that
 1 ($c \in$ CLAUSES ($featureSet$)) **and**
 2 ($c \neq$ MAIN-CLAUSE ($featureSet$)) **and**
 4:
  (a) ((TYPE-OF-SOA (SOA-OF-CLAUSE($c, featureSet$),$featureSet$) = private-state) **or**
  (b):
   (i) ((TYPE-OF-SOA (SOA-OF-CLAUSE ($c, featureSet$), $featureSet$) = private-state-action) **and**
   (ii) (TO-BE-TREATED-AS-A-PRIVATE-STATE
    (SOA-OF-CLAUSE ($c, featureSet$), $featureSet, context$)))) **and**
 5: there does not exist a $c'$ such that
  ($c' \in$ CLAUSES($featureSet$)) **and**
  (TYPE-OF-SOA (SOA-OF-CLAUSE ($c', featureSet$), $featureSet$) $\in$
   {private-state, private-state-action}) **and**
  ($c' \in$ SUBORDINATING-CLAUSES ($c, featureSet$))

### AI.4 High-Level Functions.

TRACK-POV (a procedure)
% Interpretations and contexts are defined in section 6.
$i \leftarrow 1$

$context \leftarrow \langle\{\}, \{\}, \{\}, \text{presubjective-nonactive}\rangle$
**loop**
    **if** $\neg$SENTENCE(ITEM($text, i$)) **then**
        $context \leftarrow$ NEW-CONTEXT$'$(ITEM($text, i$), $context$)
    **else**
        $interpretation \leftarrow$ POV(FEATURES(ITEM($text,i$)),$context$)
        $context \leftarrow$ NEW-CONTEXT($interpretation, context$)
    **end if**
    $i \leftarrow i + 1$
**end loop**

NEW-CONTEXT
    ($interpretation = \langle value, character \rangle$,
    $context = \langle lastSC,\ lastActiveCharacter,\ previousSCs,\ textSituation\rangle$): context
**if** $value =$ subjective **then**
    $lsc_{new} \leftarrow$ character
    $psc_{new} \leftarrow character \cup previousSCs$
    $lac_{new} \leftarrow lastActiveCharacter$
    $ts_{new} \leftarrow$ continuing-subjective
**else**
    **if** $character \neq \{\}$ **then**
        $lac_{new} \leftarrow$ character
    **else**
        $lac_{new} \leftarrow lastActiveCharacter$
    **end if**
    $lsc_{new} \leftarrow lastSC$
    $psc_{new} \leftarrow previousSCs$
    **if** $character \neq \{\}$ **and** $textSituation =$ presubjective-nonactive **then**
        $ts_{new} \leftarrow$ presubjective-active
    **else if** $character \neq \{\}$ **and** $textSituation \in \{$postsubjective-nonactive, broken-subjective$\}$ **then**
        $ts_{new} \leftarrow$ postsubjective-active
    **else if** $character = \{\}$ **and** $textSituation =$ broken-subjective **then**
        $ts_{new} \leftarrow$ postsubjective-nonactive
    **else if** $textSituation =$ continuing-subjective **then**
        $ts_{new} \leftarrow$ interrupted-subjective
    **else** $ts_{new} \leftarrow textSituation$
    **end if**
**end if**
**return** $\langle lsc_{new}, lac_{new}, psc_{new}, ts_{new}\rangle$

NEW-CONTEXT$'$
    ($break$,
    $context = \langle lastSC,\ lastActiveCharacter,\ previousSCs,\ textSituation\rangle$): context
**if** $break =$ scene-break **then**
    $ts_{new} \leftarrow$ presubjective-nonactive
**else if** $textSituation =$ presubjective-active **then**
    $ts_{new} \leftarrow$ presubjective-nonactive
**else if** $textSituation =$ continuing-subjective **then**
    $ts_{new} \leftarrow$ broken-subjective
**else if** $textSituation =$ interrupted-subjective **then**
    $ts_{new} \leftarrow$ postsubjective-nonactive

```
    else if textSituation = postsubjective-active then
        ts_new ← postsubjective-nonactive
    else ts_new ← textSituation
    end if
    return ⟨lastSC, lastActiveCharacter, previousSCs, ts_new⟩
```

POV(*featureSet, context*): *Interpretation*
**if** SENTENCE-IS-SUBJECTIVE (*featureSet, context*) **then**
    **return** ⟨ subjective, IDENTIFY-SC (*featureSet, context*) ⟩
**else**
    **return** ⟨ objective, ACTIVE-CHARACTER-OF (*featureSet, context*) ⟩
**end if**

SENTENCE-IS-SUBJECTIVE(*featureSet, context*): boolean
*soa* ← CHOSEN-STATE-OF-AFFAIRS(*featureSet, context*)
**return**
    1 (NARRATIVE-PARENTHETICAL(*featureSet*)) **or**
    2 (SUBJECTIVE-ELEMENTS(*featureSet, context*) ≠ {})) **or**
    3 (TYPE-OF-SOA(*soa, featureSet*) = private-state) **or**
    4:
        (a) ((TYPE-OF-SOA(*soa, featureSet*) = private-state-action) **and**
        (b)  (TO-BE-TREATED-AS-A-PRIVATE-STATE(*soa, featureSet, context*)) **or**
    5:
        (a) ((TYPE-OF-SOA(*soa, featureSet*) = nonprivate-state) **and**
        (b)  (TEXT-SITUATION-OF(*context*) = continuing-subjective))

SUBJECTIVE-ELEMENTS(*featureSet, context*): set of potential subjective elements
% As specified in section AI.1, the potential subjective element categories and the
% text situations with which they are associated are not listed in this paper.
**return**
    {*pse* | *pse* ∈ POTENTIAL-SUBJECTIVE-ELEMENTS (*featureSet*) **and**
        *pse* is associated with TEXT-SITUATION-OF(*context*)}

IDENTIFY-SC (*featureSet, context*): set of characters
**if** IDENTIFY-SC-FROM-THE-SENTENCE (*featureSet, context*) ≠ {} **then**
    **return** IDENTIFY-SC-FROM-THE-SENTENCE (*featureSet, context*)
**else if** LAST-SC-IS-AN-EXPECTED-SC (*context*) **and**
    LAST-ACTIVE-CHARACTER-IS-AN-EXPECTED-SC (*context*) **then**
    **return** CHOOSE-AN-EXPECTED-SC (*featureSet, context*)
**else if** LAST-SC-IS-AN-EXPECTED-SC (*context*) **then**
    **return** LAST-SC-OF(*context*)
**else if** LAST-ACTIVE-CHARACTER-IS-AN-EXPECTED-SC (*context*) **then**
    **return** LAST-ACTIVE-CHARACTER-OF(*context*)
**else**
    **return** {}
**end if**

IDENTIFY-SC-FROM-THE-SENTENCE (*featureSet, context*): set of characters
*soa* ← CHOSEN-STATE-OF-AFFAIRS (*featureSet, context*)
**if** NARRATIVE-PARENTHETICAL(*featureSet*) **then**
    **return** SUBJECT-OF-NARRATIVE-PARENTHETICAL(*featureSet*)
**else if**

    1 (EXPERIENCER-OR-ACTOR-OF(*soa, featureSet*) ≠ {}) **and**
    2 (SUBJECTIVE-ELEMENTS-TO-CONSIDER(*soa, featureSet, context*) = {}) **and**
    3:
        (a) ((TYPE-OF-SOA(*soa, featureSet*) = private-state) **or**
        (b):
            (i) ((TYPE-OF-SOA(*soa, featureSet*) = private-state-action) **and**
            (ii) (TO-BE-TREATED-AS-A-PRIVATE-STATE(*soa, featureSet, context*)))) **and**
    4:
        (a) ((TEXT-SITUATION-OF(*context*) ≠ continuing-subjective) **or**
        (b):
            (i) ((TEXT-SITUATION-OF(*context*) = continuing-subjective) **and**
            (ii):
                (1) ((EXPERIENCER-OR-ACTOR-OF(*soa*) ⊂ LAST-SC-OF(*context*) **or**
                (2) (EXPERIENCER-OR-ACTOR-OF(*soa*) ⊃ LAST-SC-OF(*context*))))
    **then return** EXPERIENCER-OR-ACTOR-OF(*soa* )
**else return** {}
**end if**

LAST-SC-IS-AN-EXPECTED-SC(*context*): boolean
**If** TEXT-SITUATION-OF(*context*) ∉ {presubjective-nonactive, presubjective-active} **then**
    **return** true
**else return** false

LAST-ACTIVE-CHARACTER-IS-AN-EXPECTED-SC(*context*): boolean
**If** TEXT-SITUATION-OF(*context*) ∈ {presubjective-active, postsubjective-active} **then**
    **return** true
**else return** false
**end if**

CHOOSE-AN-EXPECTED-SC(*featureSet, context*): set of characters
**if** EXPERIENCER-OR-ACTOR-OF(CHOSEN-STATE-OF-AFFAIRS(*featureSet*), *featureSet*)
    = LAST-ACTIVE-CHARACTER-OF(*context*) **then**
    **return** LAST-SC-OF(*context*)
**else**
    **return** LAST-ACTIVE-CHARACTER-OF (*context*)
**end if**

SUBJECTIVE-ELEMENTS-TO-CONSIDER(*soa, featureSet, context*): set of potential subjective elements
**return**
    {*pse* | *pse* ∈ SUBJECTIVE-ELEMENTS (*featureSet, context*) **and**
        ¬ IS-PSE-SUBORDINATED-TO-SOA(*pse, soa, featureSet*) **and**
        TYPE-OF-PSE(*pse, featureSet*) ∉
            {habitual, comparative-'like', 'as'-followed-by-modifier, some kinds of intensifier adverbs.}}
        % See section 10 and item (6) of AI.1.

ACTIVE-CHARACTER-OF (*featureSet, context*): set of characters
*soa* ← CHOSEN-STATE-OF-AFFAIRS(*featureSet, context*)
**if** (TYPE-OF-SOA(*soa, featureSet*) = action) **and**
  (EXPERIENCER-OR-ACTOR-OF(*soa, featureSet*) ⊆ Previous-SCs-Of (context)) **and**
  (IS-IN-THE-SIMPLE-PAST(CLAUSE-OF(*soa, featureSet*), *featureSet*)) **and**
  % the remaining conjuncts of this conditional are in English, so as not to list many
  % obvious functions. IS-IN-THE-SIMPLE-PAST is a model for the ones implied by the English statements.

(The main verb phrase is not negated, is not habitual, and does not contain a
modal auxiliary verb or adverb such as those listed in Section 8.2.2)
**then return** EXPERIENCER-OR-ACTOR-OF(*soa, featureSet*)
**else return** {}
**end if**

### Appendix II: Implementation and Demonstrations

#### Implementation.

The algorithm has been implemented in two systems. Both implement functions POV, NEW-CONTEXT, and NEW-CONTEXT', and neither implements function ITEM. The system demonstrated here takes actual sentences as input, so has a component for performing function FEATURES. However, it does not truly implement FEATURES, but is successful in computing the *FeatureSet* of a sentence only for sentences that fall within its limited coverage. It requires many actual sentences of a text to be simplified.

The other version of the system queries the user for the information returned by function FEATURES, to enable the algorithm to be tested on unlimited text, without concern for problems not addressed in this work.

Both systems are implemented using the SNePS knowledge representation and reasoning system (Shapiro 1979; Shapiro and Rapaport 1987) and an ATN grammar (Shapiro 1982). The grammar of the system demonstrated here is an extension of others developed at the State University of New York at Buffalo, and includes pieces of programs written by Soon Ae Chun (Chun 1987), Zuzana Dobes, Naicong Li (Li 1987), Sandra Peters (Peters and Shapiro 1987ab; Peters, Shapiro, and Rapaport 1988), William J. Rapaport (Rapaport 1986), and Stuart C. Shapiro (Shapiro 1982, Shapiro and Rapaport 1987).

#### Demonstrations.

In the following, input to the system is preceded by a colon, comments are preceded by percent signs, and all other lines are the system's output. To save space, extraneous messages have been deleted, such as those concerned with entering and exiting the system. Sentences of quoted speech are input simply as 'Quoted_speech' followed by a discourse parenthetical, since the algorithm does not consider the contents of the quoted string.

The system is first demonstrated on the following passage; when the passage is encountered in the novel, the situation is postsubjective-nonactive and the last subjective character is Sandy and Dennys.

> Japheth looked at Sandy and Dennys anxiously. "Sun-sickness can be dangerous." He reached up to touch Dennys's cheek. Shook his head. "You're cold and clammy. Bad sign." He put his hand against his forehead. Appeared to be thinking deeply. [L'Engle, *Many Waters*, p. 24]

```
: Initialize situation to postsubj-nonactive.
The situation is now postsubj-nonactive

: Initialize last_subj_char to Sandy and Dennys.
Dennys and Sandy is the last_subj_char
```
% *The situation and last subjective character are first initialized.*

```
: Japheth looked at Sandy and Dennys anxiously.
At the beginning of this sentence:
```

```
    The situation is postsubj-nonactive
    Expected subjective character:
        Dennys and Sandy, the last_subj_char
Perc_action of Japheth treated as an action:    Actor has not been the subj_char
The sentence is not subjective
The situation is still postsubj-nonactive
```
% *The state of affairs that the algorithm considers is the perceptual action ('perc-action')*
% *that the main clause is about, a kind of private-state action.*

% *Note that even though 'anxiously' is a private-state term, it isn't considered*
% *by the system because it is being used as a manner adverbial.*

```
: Quoted_speech Japheth said.
At the beginning of this sentence:
    The situation is postsubj-nonactive
    Expected subjective character:
        Dennys and Sandy, the last_subj_char
The sentence is not subjective
The situation is still postsubj-nonactive

: He reached up to touch Dennys's cheek.
At the beginning of this sentence:
    The situation is postsubj-nonactive
    Expected subjective character:
        Dennys and Sandy, the last_subj_char
The sentence is not subjective
The situation is still postsubj-nonactive

: Shook his head.
At the beginning of this sentence:
    The situation is postsubj-nonactive
    Expected subjective character:
        Dennys and Sandy, the last_subj_char
Potential subjective element considered:
    sentence_fragment
It is a subjective element
Subjective context established by this feature:
    sentence_fragment
The subj_char is Dennys and Sandy
The situation is now continuing-subj
```
% *This utterance is a sentence fragment, a potential subjective element. In the situation*
% *in which it appears (postsubjective-nonactive), it is a subjective element.*

```
: Quoted_speech he said.
At the beginning of this sentence:
    The situation is continuing-subj
    Expected subjective character:
        Dennys and Sandy, the last_subj_char
The sentence is not subjective
Objective sentence in continuing-subj situation:
    situation is now interrupted-subj
```

```
: He put his hand against his forehead.
At the beginning of this sentence:
  The situation is interrupted-subj
  Expected subjective character:
      Dennys and Sandy, the last_subj_char
The sentence is not subjective
The situation is still interrupted-subj

: Appeared to be thinking deeply.
At the beginning of this sentence:
  The situation is interrupted-subj
  Expected subjective character:
      Dennys and Sandy, the last_subj_char
Potential subjective elements considered:
  sentence_fragment
  progressive
  seeming_verb
Of these, the following are subjective elements:
  sentence_fragment
  seeming_verb
Subjective context established by these features:
  sentence_fragment
  seeming_verb
The subj_char is Dennys and Sandy
The situation is now continuing-subj
```
% *Even though the sentence is a private-state sentence that is not in the*
% *continuing-subjective situation, the system identifies the subjective*
% *character to be the last subjective character, because non-subordinated*
% *subjective elements appear (note that the progressive aspect is not a*
% *subjective element in this situation).*

The next demonstration is on a slightly simplified version of (25.2)–(25.3). (When it is encountered, the last subjective character is the girl.)

> How could the poor thing have married him?
>     Johnnie Martin could not believe that he was seeing that old bag's black eyes sparkling with disgust.

```
: Initialize situation to continuing-subj.
The situation is now continuing-subj

: Initialize last_subj_char to the girl.
the girl is the last_subj_char

: How could the poor thing have married him?
At the beginning of this sentence:
  The situation is continuing-subj
  Expected subjective character:
      the girl, the last_subj_char
Potential subjective elements considered:
  question
```

```
    past_perfective
    eval_adjective
All of these are subjective elements
Subjective context continued by these features:
    question
    eval_adjective
    past_perfective
The subj_char is the girl
The situation is still continuing-subj
% The system's abbreviation for an evaluative adjective is 'eval_adjective'.
% The evaluative adjective in this sentence is 'poor'.

: Paragraph.
Before the paragraph break:
    The situation is continuing-subj
    Expected subjective character:
        the girl, the last_subj_char
After the paragraph break:
     The situation is broken-subj
    The last_subj_char is still an expected subjective character
% 'Paragraph' indicates that a paragraph break occurs at this point.

: Johnnie Martin could not believe that he was seeing the old bag's black
    eyes sparkling with disgust.
At the beginning of this sentence:
    The situation is broken-subj
    Expected subjective character:
        the girl, the last_subj_char
Potential subjective elements not considered:
    percept_term
    attitude_noun
Subjective context established by this feature:
    private_state of |Johnnie Martin|
The subj_char is |Johnnie Martin|
The situation is now continuing-subj
% The system does not consider the perceptual term ('sparkling') or the attitude noun
% ('old bag') because they are subordinated to the private-state term.
```

In the following passage, competition arises that is resolved in favor of the last subjective character:

> Newt had always missed having a father, but the fact that Sean spoke so coldly of his put the matter in a different light. Perhaps he was not so unlucky, after all.
>
> He was riding around the herd when Jake Spoon trotted past on his way to Lonesome Dove.
>
> "Going to town, Jake?" Newt asked.
>
> "Yes, I think I will," Jake said. He didn't stop to pass the time; in a second, he was out of sight in the shadows. It made Newt's spirits fall a little, for Jake had seldom said two words to him since he came back. [McMurtry, *Lonesome Dove*, p. 200]

The system will be demonstrated on a slightly modified version of the critical part of
this passage (Newt is the last subjective character at the beginning):

> Newt was riding around the herd when Jake Spoon went by on his way
> to Lonesome Dove.
> "Going to town, Jake?" Newt asked.
> "Yes, I think I will," Jake said. He didn't stop to pass the time. In a second
> he was out of sight in the shadows.

```
: Previous_subj_char Jake.
Jake has been the subj_char
```
% *First, the system has to be informed that Jake has been the subjective character.*

```
: Initialize situation to broken-subj.
The situation is now broken-subj

: Initialize last_subj_char Newt.
Newt is the last_subj_char

: Newt was riding around the herd when Jake went by on his way to Lonesome Dove.
At the beginning of this sentence:
  The situation is broken-subj
  Expected subjective character:
      Newt, the last_subj_char
Potential subjective element considered:
  progressive
It is not a subjective element
Newt is the active_char of this sentence
The sentence is not subjective
Sentence with an active_char in broken-subj situation:
  situation is now postsubj-active
```
% *The progressive aspect is a subjective element only in the continuing-subjective situation.*

```
: Paragraph.
Before the paragraph break:
  The situation is postsubj-active
  Expected subjective characters:
      Newt, the last_subj_char
      Newt, the last_active_char
After the paragraph break:
  The situation is postsubj-nonactive
  The last_active_char is no longer an expected subjective character
  The last_subj_char is still an expected subjective character

: Quoted_speech Newt asked.
At the beginning of this sentence:
  The situation is postsubj-nonactive
  Expected subjective character:
      Newt, the last_subj_char
Newt is the active_char of this sentence
The sentence is not subjective
```

```
Sentence with an active_char in postsubj-nonactive situation:
  situation is now postsubj-active

: Paragraph.
Before the paragraph break:
  The situation is postsubj-active
  Expected subjective characters:
      Newt, the last_subj_char
      Newt, the last_active_char
After the paragraph break:
  The situation is postsubj-nonactive
  The last_active_char is no longer an expected subjective character
  The last_subj_char is still an expected subjective character

: Quoted_speech Jake said.
At the beginning of this sentence:
  The situation is postsubj-nonactive
  Expected subjective character:
      Newt, the last_subj_char
Jake is the active_char of this sentence
The sentence is not subjective
Sentence with an active_char in postsubj-nonactive situation:
  situation is now postsubj-active

: He did not stop to pass the time.
At the beginning of this sentence:
  The situation is postsubj-active
  Expected subjective characters:
      Newt, the last_subj_char
      Jake, the last_active_char
The sentence is not subjective
The situation is still postsubj-active

: In a second he was out of sight in the shadows.
At the beginning of this sentence:
  The situation is postsubj-active
  Expected subjective characters:
      Newt, the last_subj_char
      Jake, the last_active_char
Potential subjective element considered:
  percept_term
It is a subjective element
Competition between the last_subj_char and the last_active_char
  Choosing the last_subj_char because the sentence is about the last_active_char
Subjective context established by this feature:
  percept_term
The subj_char is Newt
The situation is now continuing-subj
```
*% The percept term in this sentence is 'sight'.*

**Appendix III: A Test of the Algorithm**

**AIII.1 Introduction.**

To give an idea of the success rate of the algorithm, this appendix presents the results of a test of the algorithm on 450 sentential input items (exclusive of paragraph and scene breaks) from each of two novels, *Lonesome Dove* by Larry McMurtry and *The Magic of the Glits* by Carole S. Adler. *Lonesome Dove* is an adult novel that has many subjective characters, and *The Magic of the Glits* is a childrens' novel that has one main subjective character. The input items are those of the complete sentences of every fifth page of these novels, starting in *Lonesome Dove* with page 176 and ending with page 236 (13 pages total), and starting in *The Magic of the Glits* with page 1 and ending with page 86 (18 pages total). (For each book, the first part of an additional page was used to make the number of input items exactly equal to 450.) Page 176 in *Lonesome Dove* is the beginning of a chapter in the middle of the novel. The reason why I started in the middle of the novel is that earlier pages were considered during the development of the algorithm.

The system used in this study implements only POV, NEW-CONTEXT, and NEW-CONTEXT'. The results of ITEM and FEATURES are supplied by me. What is being tested in this study is POV—NEW-CONTEXT and NEW-CONTEXT' are not in question, since their mappings follow from the definitions of an interpretation and of a context and its components. What POV is judged against is function H, which is based on my judgements. It maps a sentence into the correct interpretation of the sentence:

$$\text{H} : sentence \rightarrow interpretation.$$

Sometimes, either a subjective or objective interpretation of a sentence would be reasonable. For these sentences, I accept the algorithm's interpretation—assuming that $s$ is such a sentence, and $c$ is the context in which $s$ appears, I take H($s$) to be POV(FEATURES($s$), $c$). There are fewer than 20 such sentences in the passages from *Lonesome Dove* and fewer than 30 in the passages from *The Magic of the Glits*. One goal of our current and planned psychological experiments is to identify the types of situations in which there are significant individual differences among subjects' interpretations.

We will also distinguish between *primary* and *secondary* errors, primary errors being the more severe. The algorithm's interpretation of a sentence is a *primary error* when its interpretation is incorrect given the *actual context* of the sentence, which is computed from the correct interpretations of all previous sentences. A *secondary error* is one that results only from previous errors. In this case, the algorithm's interpretation is correct given the actual context, but incorrect given the context computed by the algorithm. The definition of the context computed by the algorithm is given above in section 6.2; the definition of the actual context of the $i^{th}$ input item, $ac_i$, is the same, except that H takes the place of POV:

$$ac_i = \begin{cases} \langle\{\},\{\},\{\},\text{presubjective-nonactive}\rangle & \text{if } i=1 \\ \text{NEW-CONTEXT'}(\text{ITEM }(t,i-1), c_{i-1}) & \text{if } i>1 \text{ \& } \neg\text{SENTENCE }(\text{ITEM }(t,i-1)) \\ \text{NEW-CONTEXT}(\text{H}(\text{ITEM }(t,i-1)), ac_{i-1}) & \text{if } i>1 \text{ \& } \text{SENTENCE }(\text{ITEM }(t,i-1)) \end{cases}$$

**AIII.2 Results.**

**Lonesome Dove**
We now present the results of the study, beginning with *Lonesome Dove*. Out of the 450 input items, the algorithm committed 27 primary errors (6%) and 28 secondary errors (6%). We will first give a breakdown of the primary errors according to interpretation in table 4, and then one according to point-of-view operation in table 5. Note that many of the input items, 125 of them (28%), are simple items of quoted speech (i.e., they do not have potential subjective elements in the discourse parenthetical, or subordinated clauses outside the quoted string that have private-state terms, private-state-action terms, or potential subjective elements).

Consider table 4. The first row, for example, should be understood as follows: out of the 271 actual subjective sentences, the algorithm committed 20 primary errors. It interpreted 13 subjective sentences to be objective, and 7 to be the subjective sentence of the wrong subjective character.

| Interpretation | Actual Instances | Primary Errors | Incorrect Interpretations |
|---|---|---|---|
| $\langle$subjective,$x\rangle$ | 271/450 (60%) | 20/271 (7%) | 13 objective<br>7 $\langle$subjective,$y\rangle$, $y \neq x$ |
| objective | 179/450 (40%) | 7/179 (4%) | 7 $\langle$subjective,$x\rangle$ |
| objective, other than simple quoted speech | 54/450 (12%) | 7/54 (13%) | 7 $\langle$subjective,$x\rangle$ |

**Table 4**
Results for *Lonesome Dove* by interpretation.

Now consider table 5. The first row, for example, should be understood as follows: out of the 215 items that actually continue a character's point of view, the algorithm committed 11 primary errors. It interpreted 1 of them to be an initiation and 10 to be objective. Notice that the last column of the row for initiations includes an initiation. This means that for one actual initiation, the algorithm was correct that a character's point of view was initiated, but incorrect as to the identity of that character.

| Point-of-View Operation | Actual Instances | Primary Errors | Incorrect Interpretations |
|---|---|---|---|
| continuation | 215/450 (48%) | 11/215 (5%) | 1 initiation<br>10 objective |
| resumption | 20/450 (4%) | 0/20 (0%) | — |
| initiation | 36/450 (8%) | 9/36 (25%) | 5 resumptions<br>1 initiation<br>3 objective |
| objective | 179/450 (40%) | 7/179 (4%) | 4 continuations<br>3 resumptions |
| objective, other than simple quoted speech | 54/450 (12%) | 7/54 (13%) | 4 continuations<br>3 resumptions |

**Table 5**
Results for *Lonesome Dove* by point-of-view operation.

### The Magic of the Glits

In *The Magic of the Glits*, out of the 450 input items, the algorithm committed 34 primary errors (8%) and 21 secondary errors (5%). There are 228 items that are simple quoted speech (51%). Tables 6 and 7 present the kinds of results for this novel that were given above in tables 4 and 5 for *Lonesome Dove*.

A particular weakness of the algorithm on novels such as *The Magic of the Glits* is responsible for a number of the primary errors given in the last two rows of tables 6 and 7. The salient feature of the novel is that it has two main characters, Jeremy and Lynette, but primarily takes the psychological point of view only of Jeremy. The problem with the algorithm is that very minor subjective contexts of Lynette affect the interpretations of later sentences about her private-state actions—they are treated as private-state sentences, while they should be treated as action sentences.

| Interpretation | Actual Instances | Primary Errors | Incorrect Interpretations |
| --- | --- | --- | --- |
| $\langle$subjective,$x\rangle$ | 125/450 (28%) | 12/125 (10%) | 10 objective<br>2 $\langle$subjective,$y\rangle$, $y \neq x$ |
| objective | 325/450 (72%) | 22/325 (7%) | 22 $\langle$subjective,$x\rangle$ |
| objective, other than simple quoted speech | 97/450 (22%) | 22/97 (23%) | 22 $\langle$subjective,$x\rangle$ |

**Table 6**
Results for *The Magic of the Glits* by interpretation.

| Point-of-View Operation | Actual Instances | Primary Errors | Incorrect Interpretations |
| --- | --- | --- | --- |
| continuation | 79/450 (18%) | 4/79 (5%) | 4 objective |
| resumption | 41/450 (9%) | 7/41 (17%) | 2 initiations<br>5 objective |
| initiation | 5/450 (1%) | 1/5 (20%) | 1 objective |
| objective | 325/450 (72%) | 22/325 (7%) | 4 continuations<br>9 resumptions<br>9 initiations |
| objective, other than simple quoted speech | 97/450 (22%) | 22/97 (23%) | 4 continuations<br>9 resumptions<br>9 initiations |

**Table 7**
Results for *The Magic of the Glits* by point-of-view operation.

When a novel primarily takes an *external view* of a character (Uspensky 1973), descriptions of behavior, quoted speech, and private-state-action sentences are the primary means employed to communicate things about his or her consciousness. However, if a character appears often enough, some reports of that character's private states are bound to appear to explain his or her actions. According to the algorithm as presented in this paper, once a character is the SC of any subjective sentence, all later private-state actions of that character are treated as private states rather than as actions (see section 11). As suggested in Wiebe 1990, however, this criterion is too weak—what should be required for a private-state action to be treated as a private state is that there be a previous subjective context of the actor that is "significant". Some possible, but perhaps arbitrary, definitions of a "significant" subjective context are that it contain a represented thought, that it contain a potential subjective element that is indeed subjective, or that it be at least two input items in length. The only reason that one of these heuristics has not been incorporated into the algorithm is that an examination of texts broad enough to choose among them (or to suggest another) has not yet been performed. For *The Magic of the Glits*, any such heuristic would suffice. In the entire novel (not only in the passages tested), there are a total of 10 subjective sentences attributed to Lynette. All are private-state reports, none has a subjective element, and no two of them appear together. In addition, 5 of them include some description of Lynette's behavior, and, of the remaining 5, 4 are the subjective sentence of Jeremy and Lynette together, not of Lynette alone. If one of the above heuristics were employed, then there would be 12 fewer primary errors and 14 fewer secondary errors. Whether or not one of these heuristics is employed does not affect the results given above for *Lonesome Dove*.

Assuming one of the above heuristics, the results for *The Magic of the Glits* would be as follows. Out of the 450 input items, the algorithm would commit 22 primary errors (5%) and 7 secondary errors (2%). Table 8 shows the primary errors broken down according to point-of-view operation, assuming that one of the heuristics discussed above is employed. Notice that the differences between tables 7 and 8 appear in the last two rows.

| Point-of-View Operation | Actual Instances | Primary Errors | Incorrect Interpretations |
|---|---|---|---|
| continuation | 79/450 (18%) | 4/79 (5%) | 4 objective |
| resumption | 41/450 (9%) | 7/41 (17%) | 2 initiations<br>5 objective |
| initiation | 5/450 (1%) | 1/5 (20%) | 1 objective |
| objective | 325/450 (72%) | 10/325 (3%) | 1 continuation<br>9 resumptions |
| objective, other than simple quoted speech | 97/450 (22%) | 10/97 (10%) | 1 continuation<br>9 resumptions |

**Table 8**
Revised results for *The Magic of the Glits* by point-of-view operation.